\documentclass[pdflatex,sn-nature]{sn-jnl}
\setcitestyle{super,open={},close={}}

\usepackage{graphicx}%
\graphicspath{{figures/}}
\usepackage{multirow}%
\usepackage{amsmath,amssymb,amsfonts}%
\usepackage{amsthm}%
\usepackage{mathrsfs}%
\usepackage[title]{appendix}%
\usepackage{xcolor}%
\usepackage{overpic}%
\usepackage{colortbl}%
\usepackage{textcomp}%
\usepackage{manyfoot}%
\usepackage{booktabs}%
\usepackage{array}%
\usepackage{tabularx}%
\usepackage{algorithm}%
\usepackage{algorithmicx}%
\usepackage{algpseudocode}%
\usepackage{listings}%
\usepackage{placeins}
\usepackage{needspace}
\makeatletter\@secpenalty=\z@\makeatother
\makeatletter\let\ml@section\section\renewcommand{\section}{\Needspace{4\baselineskip}\ml@section}\let\ml@subsection\subsection\renewcommand{\subsection}{\Needspace{6\baselineskip}\ml@subsection}\makeatother
\usepackage{caption}
\usepackage{ml-arxiv-readable}
\usepackage{rotating}

\makeatletter
\def\sectionfont{\reset@font\sffamily\fontsize{13pt}{16.2pt}\bfseries\selectfont\color{mlInk}\raggedright\boldmath}
\def\subsectionfont{\reset@font\sffamily\fontsize{11.4pt}{14.3pt}\bfseries\selectfont\color{mlInk}\raggedright\boldmath}
\def\subsubsectionfont{\reset@font\rmfamily\fontsize{10pt}{13pt}\bfseries\selectfont\color{mlInk}\boldmath}
\def\toclevel@subparagraph{1}
\renewcommand\section{\@startsection{section}{1}{\z@}{-18pt \@plus -3pt \@minus -2pt}{8pt}{\sectionfont}}
\renewcommand\subsection{\@startsection{subsection}{2}{\z@}{-13pt \@plus -2pt \@minus -2pt}{5pt}{\subsectionfont}}
\renewcommand\subsubsection{\@startsection{subsubsection}{3}{\z@}{-8pt \@plus -2pt \@minus -1pt}{-0.65em}{\subsubsectionfont}}
\makeatother

\definecolor{paperFrontBackground}{HTML}{F1F4F7}
\mlfrontpagebreakfalse
\makeatletter
\long\def\abstract#1{\def\@abstract{{\abstractfont #1\par}}}
\def\keywords#1{\def\@keywords{{\keywordfont\textbf{Keywords}\hspace{0.65em}#1\par}}}
\renewcommand{\@maketitle}{%
  \newpage
  \begingroup
    \setlength{\parindent}{0pt}%
    \setlength{\parskip}{0pt}%
    \begin{tcolorbox}[
      enhanced,width=\textwidth,frame hidden,
      colback=paperFrontBackground,boxrule=0pt,boxsep=0pt,
      arc=10pt,outer arc=10pt,
      left=14pt,right=14pt,top=16pt,bottom=16pt,
      before skip=0pt,after skip=0pt]
      {\Titlefont\@title\par}%
      \ifx\@subtitle\empty\else
        \vspace{5pt}{\SubTitlefont\@subtitle\par}%
      \fi
      \ifnum\aucount>0
        \vspace{9pt}%
        {\def\Authorfont{\reset@font\rmfamily\fontsize{10.5pt}{13pt}\bfseries\selectfont\color{mlInk}\raggedright}%
         \global\punctcount\aucount\artauthors\par}%
        \vspace{7pt}%
        {\addressfont\auaddress\par}%
        \ifequalcont
          \vspace{4pt}{\contactfont $^{\dagger}$\@equalconttext\par}%
        \fi
        \ifpresentaddress
          \vspace{4pt}{\contactfont\@presentaddresstext\par}%
        \fi
      \fi
      \vspace{15pt}%
      \printabstract
      \ifnum\emailcnt>0\relax
        \vspace{13pt}%
        {\contactfont\textsuperscript{*}\textbf{Corresponding authors:}\hspace{0.45em}Munyoung Chang (\href{mailto:caduent@gmail.com}{caduent@gmail.com}); Sungroh Yoon (\href{mailto:sryoon@snu.ac.kr}{sryoon@snu.ac.kr})\par}%
      \fi
      \ifx\@keywords\empty\else\vspace{7pt}\printkeywords\fi
    \end{tcolorbox}
  \endgroup
  \ifmlfrontpagebreak\clearpage\else\vspace{19pt}\fi
}
\makeatother

\theoremstyle{thmstyleone}%

\theoremstyle{thmstyletwo}%

\theoremstyle{thmstylethree}%

\newcommand{\reasonregion}{reasoning-related region}

\newcolumntype{L}[1]{>{\raggedright\arraybackslash}p{#1}}
\newsavebox{\EDbox}
\newsavebox{\EDboxB}

\begin{document}

\title[]{Transferring an analysis of human brain organization to language models}

\author[1]{\fnm{Donghyeok} \sur{Lee}}

\author[2]{\fnm{Sangwon} \sur{Yu}}

\author[3]{\fnm{Sangkwon} \sur{Park}}

\author[4,5]{\fnm{Hye Youn} \sur{Park}}

\author*[6,7,8]{\fnm{Munyoung} \sur{Chang}}\email{caduent@gmail.com}
\equalcont{These authors contributed equally as corresponding authors.}

\author*[1,3]{\fnm{Sungroh} \sur{Yoon}}\email{sryoon@snu.ac.kr}
\equalcont{These authors contributed equally as corresponding authors.}

\affil[1]{\orgdiv{Interdisciplinary Program in Artificial Intelligence}, \orgname{Seoul National University}, \orgaddress{\city{Seoul}, \postcode{08826}, \country{Republic of Korea}}}

\affil[2]{\orgname{Upstage AI}, \orgaddress{\city{Seoul}, \postcode{06130}, \country{Republic of Korea}}}

\affil[3]{\orgdiv{Department of Electrical and Computer Engineering}, \orgname{Seoul National University}, \orgaddress{\city{Seoul}, \postcode{08826}, \country{Republic of Korea}}}

\affil[4]{\orgdiv{Department of Psychiatry}, \orgname{Seoul National University Bundang Hospital}, \orgaddress{\city{Seongnam}, \country{Republic of Korea}}}

\affil[5]{\orgdiv{Department of Psychiatry}, \orgname{Seoul National University College of Medicine}, \orgaddress{\city{Seoul}, \country{Republic of Korea}}}

\affil[6]{\orgdiv{Department of Otorhinolaryngology-Head \& Neck Surgery}, \orgname{Seoul National University Bundang Hospital}, \orgaddress{\city{Seongnam}, \country{Republic of Korea}}}

\affil[7]{\orgdiv{Department of Otorhinolaryngology-Head \& Neck Surgery}, \orgname{Seoul National University College of Medicine}, \orgaddress{\city{Seoul}, \country{Republic of Korea}}}

\affil[8]{\orgdiv{Education and Research Program for Future ICT Pioneers}, \orgname{Seoul National University}, \orgaddress{\city{Seoul}, \country{Republic of Korea}}}

\abstract{During human cortical development, functional systems become increasingly differentiated.
We asked whether improved reasoning in language models involves a comparable regional reorganization.
Applying diffusion-map embedding, used to characterize cortical organization, to matched base and reasoning-trained models from three families, we found broadly preserved gradient profiles with localized changes within a contiguous layer span.
Grafting weights from these spans into instruction-tuned hosts improved mathematical reasoning on average, with gains rising with region overlap, while retaining more previously solved problems than grafts elsewhere.
Across training, the span formed gradually during the early rise in performance, and grafts from intermediate stages helped.
Grafts shifted most hosts partly towards their donors' behaviour and sometimes improved on problems the donors failed; the same analysis located a transferable span in a medical model.
Together, these findings identify a reasoning-related region whose weights improve reasoning with low interference in the host's existing mathematical competence.}


\maketitle

\begin{figure}[t]
\centering
\includegraphics[width=\textwidth]{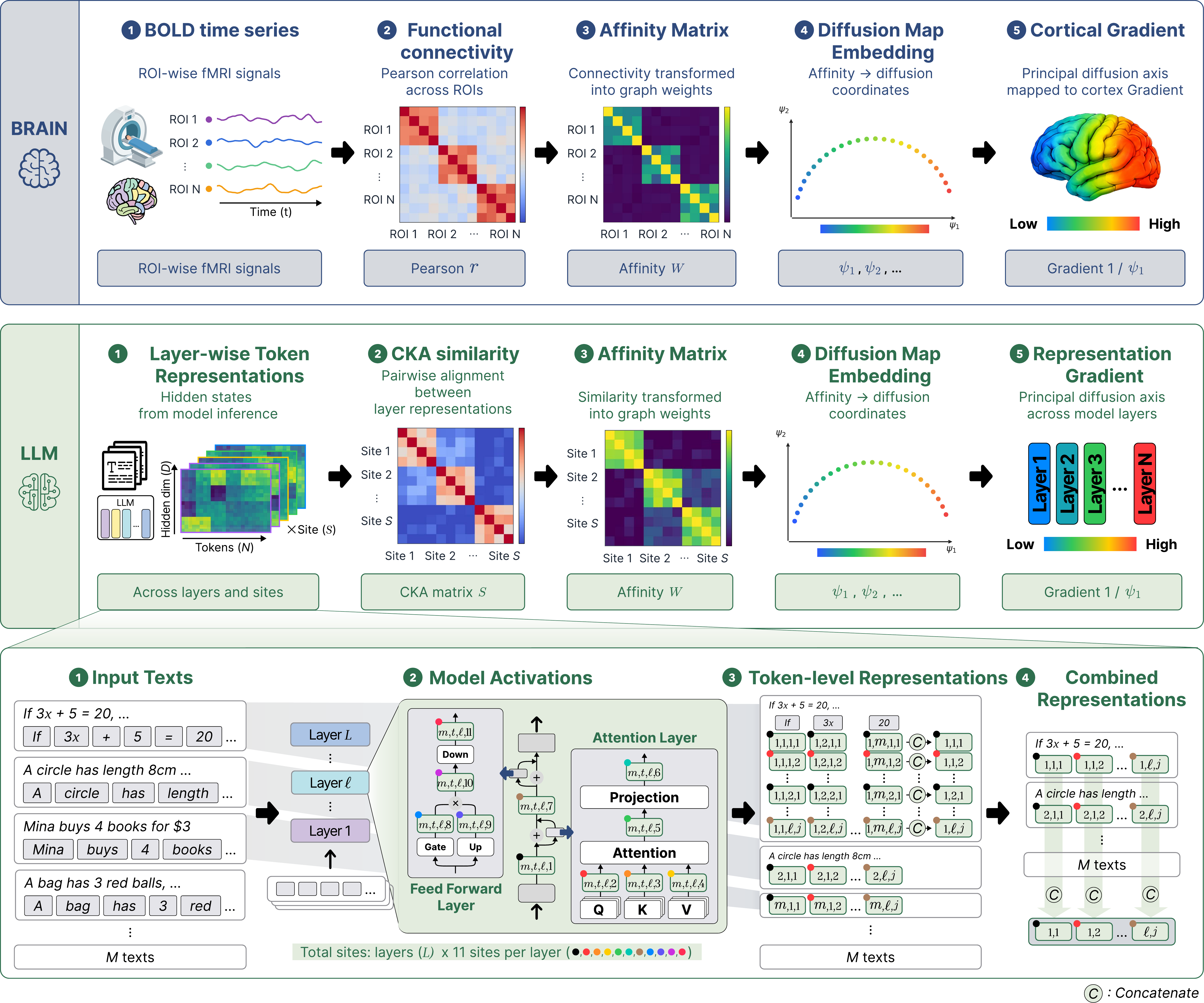}
\caption{\textbf{Transferring an analysis of cortical organization to language models.}
The top row converts fMRI/BOLD signals into functional connectivity and applies diffusion-map embedding to obtain its principal gradient. The middle row applies the same sequence to language models: CKA converts sampled activations into a representational-similarity matrix, the model-side analogue of cortical functional connectivity, and diffusion-map embedding yields a principal gradient of representational similarity. The bottom row shows activation sampling: each representation node pairs one transformer layer with one of 11 retained sites in its attention and feed-forward computations. At each node, activation vectors from all valid tokens across the input texts are stacked into a matrix; the shared row order aligns observations across nodes for pairwise CKA.}
\label{fig:overview}
\end{figure}

Reasoning abilities develop throughout childhood and adolescence, alongside changes in the functional organization of the cerebral cortex~\cite{wendelken2017,sydnor2021}.
Cortical gradient analysis describes this organization by arranging brain regions along continuous axes according to the similarity of their patterns of functional connectivity~\cite{margulies2016,huntenburg2018}.
Developmental studies using this approach have revealed a shift from a principal gradient dominated by sensory and motor systems to one that distinguishes association cortex from primary sensory and motor regions~\cite{dong2021}.
These findings prompted a question: \emph{could the same gradient analysis reveal changes in the internal organization of language models as their reasoning abilities improve?}

Pretrained language models already support a broad range of language tasks, and further training can strengthen their reasoning.
We compared pretrained base models with reasoning-trained counterparts derived from the same pretrained weights to examine how this further training changed their internal organization~\cite{deepseekr1,ministral3}.

\Needspace{6\baselineskip}
To compare these models using the cortical analysis, we needed to define which representations to examine and how to measure the relationships between them.
We sampled activation vectors from attention and feed-forward computations within each layer and measured similarities between these representations using centered kernel alignment (CKA)~\cite{kornblith2019} (Fig.~\ref{fig:overview}).
The resulting similarity matrix served as the model counterpart of cortical functional connectivity.
Diffusion-map embedding then yielded a principal gradient describing how representations related to one another across the model.

Across six model pairs spanning three families, aligned gradient profiles remained broadly similar, with localized changes detected within limited spans of neighbouring layers.
These localized changes prompted a further question: could weights from the identified spans improve reasoning in another model?
We tested this possibility by grafting them into instruction-tuned hosts of the same architecture, without further training.
These grafts improved mathematical performance on average while preserving more previously solved problems than grafts elsewhere.
We then followed a region's formation during training, examined the relationship between graft gains and donor behaviour, and extended the analysis to a medical model.
Together, these analyses connect localized representational changes to the effects of exchanging weights between models.

\FloatBarrier
\Needspace{8\baselineskip}
\section{Results}\label{sec:results}
\subsection{Localized changes in model organization}\label{sec:region}

\subsubsection{Gradient profiles were broadly preserved after reasoning training}
\begin{figure}[!t]
\centering
\begin{overpic}[width=\textwidth]{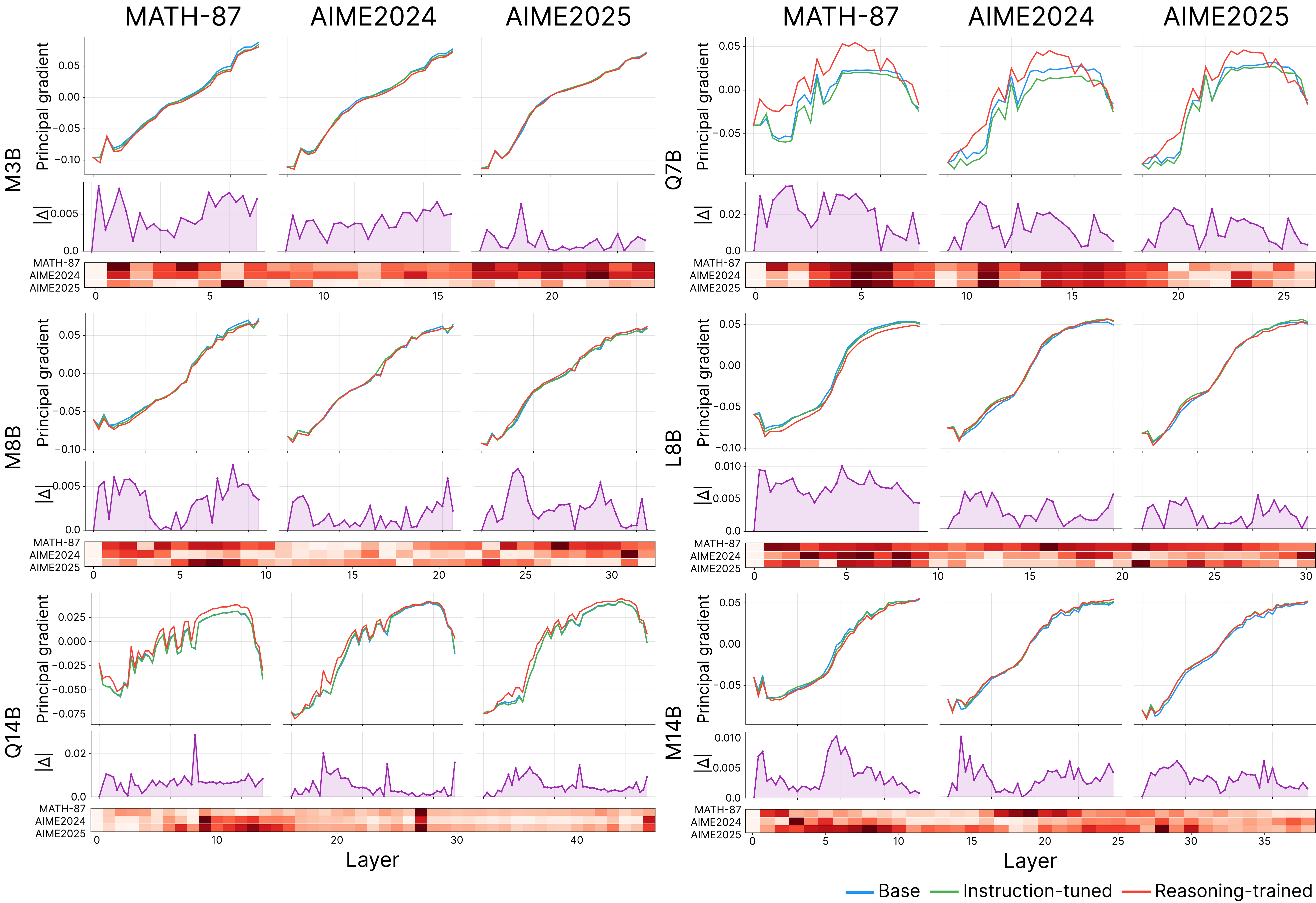}
\put(0,66){\textbf{\sffamily a}}\put(50.5,66){\textbf{\sffamily b}}
\put(0,46){\textbf{\sffamily c}}\put(50.5,46){\textbf{\sffamily d}}
\put(0,24){\textbf{\sffamily e}}\put(50.5,24){\textbf{\sffamily f}}
\end{overpic}\\[4pt]
{\footnotesize
\setlength{\tabcolsep}{8pt}
\begin{tabular}{lrrrr}
\multicolumn{5}{l}{\textbf{\sffamily g}}\\[2pt]
\toprule
\rowcolor{gray!15}Model set & $L$ & Region layers & Width & In/out ratio \\
\midrule
M3B  & 26 & 16--24 & 9 (35\%) & 1.65 \\
Q7B  & 28 & 12--18 & 7 (25\%) & 1.50 \\
M8B  & 34 & 25--31 & 7 (21\%) & 1.68 \\
L8B  & 32 & 15--22 & 8 (25\%) & 1.29 \\
Q14B & 48 & 25--31 & 7 (15\%) & 1.55 \\
M14B & 40 & 16--21 & 6 (15\%) & 2.69 \\
\bottomrule
\end{tabular}}
\caption{\textbf{Gradient displacement between base and reasoning-trained language models.}
\textbf{a}--\textbf{f}, The six matched sets of base, instruction-tuned and reasoning-trained models (M3B, Q7B, M8B, L8B, Q14B and M14B); within each, columns correspond to MATH-87, AIME~2024 and AIME~2025 inputs.
Upper, gradient~1 coordinates aligned to the base model and averaged across 11 representation sites per layer.
Lower, unsmoothed absolute base--reasoning displacement, $|\Delta_\ell|=|g^{\text{reasoning}}_\ell-g^{\text{base}}_\ell|$.
The final transformer layer is excluded.
Heatmap rows show the same displacement profiles, each normalized to its own maximum.
\textbf{g}, Span selected from each set's MATH-87 displacement profile (Methods). $L$, number of transformer layers; region layers are zero-based and inclusive; width is given as a layer count and as a percentage of $L$; in/out ratio is the mean displacement inside the span divided by the mean outside it.
Model codes are M, Ministral; Q, Qwen; and L, Llama; numerals denote parameter scale.}
\label{fig:detection}
\end{figure}

We first examined how the internal organization of language models differed before and after training to improve reasoning.
We compared six matched model pairs spanning the Ministral, Qwen and Llama families (M3B, M8B, M14B, Q7B, Q14B and L8B; the letter denotes the family and the number the parameter count in billions), using the same mathematical inputs within each pair.
The principal gradient assigned each sampled representation a coordinate according to its similarity relationships with the rest of the model (Fig.~\ref{fig:overview}; Methods).
To compare these coordinates between models, we aligned the reasoning-trained model's gradient coordinates to those of its base counterpart before averaging within layers (Methods).
Across all six pairs, the aligned gradient profiles retained similar overall shapes (Fig.~\ref{fig:detection}, upper plots).

\subsubsection{Displacement formed localized peaks that defined a span in each model}
To locate differences within these broadly similar profiles, we measured the absolute displacement at each layer,
\begin{equation}
|\Delta_\ell|=\left|g^{\mathrm{reasoning}}_\ell-g^{\mathrm{base}}_\ell\right|,
\label{eq:gradient_displacement}
\end{equation}
where $g_\ell$ denotes the aligned gradient coordinate averaged across the representations sampled within layer $\ell$.
Comparing coordinates at corresponding layers revealed differences that were less apparent from the overall profile shapes.
The displacement profiles showed localized peaks whose positions varied between model pairs and with the input dataset (Fig.~\ref{fig:detection}, lower plots).
We used the MATH-87 profiles to select one contiguous span of elevated displacement for each pair (Methods).
These spans occupied 15--35\% of the layers, with mean displacement inside each span ranging from 1.29 to 2.69 times that outside it (Fig.~\ref{fig:detection}g).
The MATH-87 spans provided fixed layer intervals for the subsequent graft experiments.

\subsubsection{Displacement did not track the magnitude of parameter change}
We then examined whether the gradient analysis identified the layers whose weights had changed most.
We measured parameter change using relative Frobenius distance~\cite{lee2019wide,zhao2026layerwise}, which summarizes how much a layer's weights differ between models relative to their magnitude in the base model (Methods).
We compared this measure with gradient displacement across layers for both base--instruction and base--reasoning transitions (Extended Data Fig.~\ref{fig:ed_weight_grad1}).
Parameter change was spread across layers with no counterpart to the displacement peaks, and across the twelve comparisons the median Spearman correlation was $\rho=-0.014$.
Gradient displacement therefore did not track the magnitude of weight change.
We next asked whether the weights within the identified spans could improve reasoning when inserted into another model.

\subsection{Reasoning gains from regional grafting}\label{sec:graft}

We tested this possibility by copying all parameters of the identified span from each reasoning-trained model into the matched instruction-tuned model, without further training (Methods).
The span identified using MATH-87 was held fixed for evaluation on MATH-87, AIME~2024 and AIME~2025.
We asked whether these grafts improved mathematical accuracy, whether the gain depended on where the weights came from, and whether the inputs used to identify the span mattered.

\subsubsection{Grafts from the identified spans improved mathematical accuracy on average}\label{sec:overlapgains}
\begin{figure}[!t]
\centering
\includegraphics[width=\textwidth]{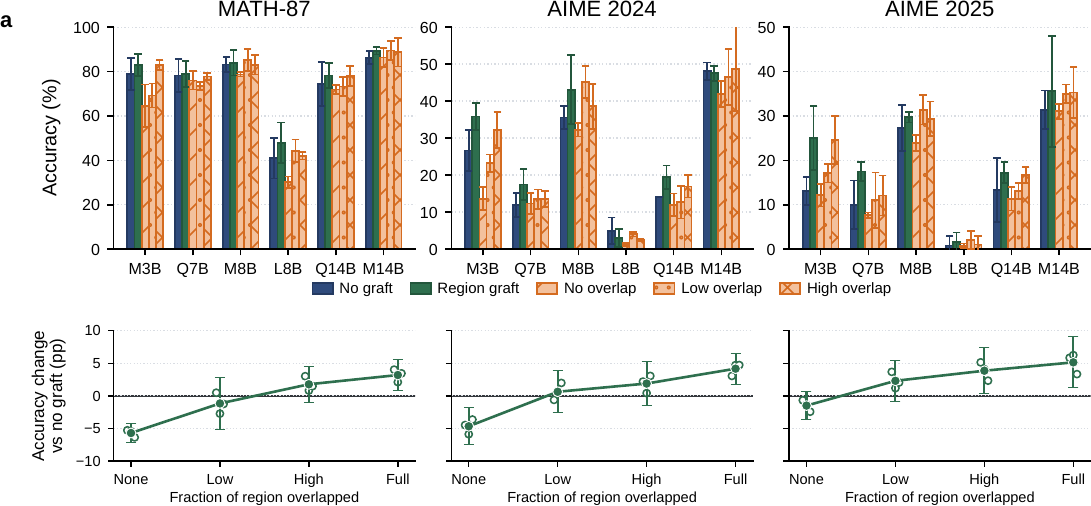}\\[3pt]
{\small
\setlength{\tabcolsep}{0pt}
\begin{tabular}{@{}L{0.115\textwidth}*{3}{>{\centering\arraybackslash}p{0.295\textwidth}}@{}}
\multicolumn{4}{@{}l}{\textbf{b}\quad Region graft minus mean across other graft positions (percentage points; mean [95\% CI])}\\
\toprule
\rowcolor{gray!15}Host & MATH-87 & AIME~2024 & AIME~2025 \\
\midrule
M3B  & $+12.5\ [3.6,21.3]$  & $+15.2\ [12.2,18.1]$ & $+8.5\ [1.9,15.1]$ \\
Q7B  & $+3.6\ [-3.7,11.0]$  & $+4.4\ [-0.6,9.4]$   & $+7.6\ [4.1,11.1]$ \\
M8B  & $+3.6\ [-2.4,9.5]$   & $+7.8\ [-2.0,17.7]$  & $+3.8\ [1.1,6.5]$ \\
L8B  & $+10.8\ [3.7,18.0]$  & $+0.9\ [-1.7,3.4]$   & $+0.6\ [-0.3,1.4]$ \\
Q14B & $+5.4\ [-2.4,13.2]$  & $+6.8\ [4.6,9.0]$    & $+5.0\ [4.4,5.6]$ \\
M14B & $+2.3\ [-0.1,4.7]$   & $+4.2\ [-0.8,9.2]$   & $+3.6\ [-8.3,15.4]$ \\
\bottomrule
\end{tabular}}
\caption{\textbf{Graft performance increases with overlap of the reasoning-related region.}
Weights from reasoning-trained models were grafted into matched instruction-tuned models (hosts), without further training.
Within each model, all grafts matched the width of its MATH-87-defined region, which was held fixed across benchmarks.
\textbf{a}, Upper plots show absolute accuracy before grafting, after the region graft and after grafts with no, low or high region overlap; lower plots show accuracy changes from the ungrafted model.
Overlap is the fraction of the region replaced: none ($0$), low ($0<\mathrm{overlap}<0.5$), high ($0.5\leq\mathrm{overlap}<1$) and full ($1$).
Positions were averaged within each model, overlap class and generation run; lower plots then weight the six models equally.
Open points show individual generation runs, and bars and filled points show their means.
\textbf{b}, Accuracy difference between the region graft and the mean across all other evaluated graft positions on the same two-layer grid.
Means and 95\% confidence intervals (whiskers in \textbf{a}, brackets in \textbf{b}) were calculated across three generation runs.
Model codes are M, Ministral; Q, Qwen; and L, Llama; numerals denote parameter scale.}
\label{fig:graft}
\end{figure}

Averaged across the six instruction-tuned models, accuracy increased relative to the ungrafted models by 3.2 percentage points on MATH-87 (95\% confidence interval, $0.8$ to $5.6$), 4.2 points on AIME~2024 ($1.8$ to $6.6$) and 5.1 points on AIME~2025 ($1.2$ to $9.1$; Fig.~\ref{fig:graft}a).
The larger mean gains on AIME extended beyond the selection problems, but were not universal: on AIME~2024, accuracy fell by 1.9 points in L8B and 0.6 points in M14B (Extended Data Table~\ref{tab:ed_other_inputs}c).

\subsubsection{Gains rose with the graft's overlap with the identified span}
We compared the identified span with grafts containing the same number of layers at other positions (Methods and Extended Data Fig.~\ref{fig:ed_stride_profiles}).
On all three benchmarks, accuracy averaged across models increased from no overlap through low and high overlap with the identified span and was greatest at full overlap, and grafts with no overlap reduced accuracy on average relative to the ungrafted models (Fig.~\ref{fig:graft}a).
The graft at the identified span also exceeded the mean accuracy of the other evaluated positions in the point estimates of all 18 model--benchmark comparisons (Fig.~\ref{fig:graft}b).
On the basis of these results, we refer to the identified span in each reasoning-trained model as its \reasonregion{}.

We also compared the region graft with equally wide grafts positioned using peaks in relative Frobenius distance or CKA similarity-profile change.
Region grafts had higher mean accuracy than both alternatives in all 18 host--benchmark comparisons (Extended Data Fig.~\ref{fig:ed_stride_profiles}).

\subsubsection{Gains extended to MATH-500 problems not used to identify the span}\label{sec:heldout}
Because each span was identified from 87 MATH-500 problems, we evaluated the region grafts on the remaining 413 problems in a single generation run.
Accuracy rose in five of six hosts, by 0.7 to 3.6 percentage points, matching the direction of change on the 87 selection problems; in M14B, where the graft left selection accuracy unchanged, held-out accuracy fell by 1.0 points (Extended Data Table~\ref{tab:ed_heldout}).
The reasoning-trained models themselves exceeded their hosts on these problems in four sets and fell below them in M8B and M14B.

\subsubsection{Spans selected from AIME problems were less effective, even on AIME}\label{sec:localizationinputs}
To examine what made an input dataset useful for identifying a span, we repeated the gradient analysis on AIME problems, on which the reasoning-trained models were less accurate than on MATH-87, keeping the model pairs unchanged.
The resulting spans overlapped the MATH-87 region in two of six model pairs for AIME~2024 and in none for AIME~2025.
When each span was evaluated on the benchmark used to identify it, against three randomly drawn grafts of the same width outside the span, the selected graft exceeded their mean in five of six models for MATH-87, three for AIME~2024 and one for AIME~2025 (Extended Data Table~\ref{tab:ed_other_inputs}a).

\subsubsection{Gains were less consistent for spans selected from non-mathematical text}
Using Wikipedia text and random word sequences as inputs, we repeated the gradient analysis and grafted weights from the resulting spans, keeping the model pairs and selection rule unchanged.
Alternative spans sometimes improved accuracy but generally yielded smaller gains than the MATH-87 region; the non-overlapping Wikipedia span nevertheless exceeded it for M14B on AIME~2024 (Extended Data Table~\ref{tab:ed_other_inputs}b,c).

\subsection{Solution length and effects on other abilities}\label{sec:knowledge}

\subsubsection{Region grafts lengthened solutions on average}
Reasoning training lengthens solutions, which come to contain intermediate calculations, checks and alternative approaches~\cite{deepseekr1}.
We therefore asked whether region grafts also transferred this tendency, comparing median response length across graft positions on the three mathematics benchmarks.
On all three benchmarks, responses were longer on average for region grafts than for non-overlapping grafts (Extended Data Fig.~\ref{fig:ed_knowledge}a).

\subsubsection{Forced continuation did not consistently reproduce the graft gains}\label{sec:forcing}
To test whether longer solutions could reproduce the gains, we made five ungrafted hosts continue whenever they stopped before either of two set lengths (Methods).
On MATH-87, region grafts had higher mean accuracy than both forced conditions in L8B, Q14B and M3B, while the shorter forced condition exceeded the region graft in M8B and M14B.
On AIME~2024 and AIME~2025, region grafts had higher mean accuracy than both forced conditions in Q14B and all three Ministral hosts, including cases where forced solutions were longer; L8B accuracy remained near zero (Extended Data Table~\ref{tab:ed_forcing}).

\subsubsection{Factual recall and language modelling changed in both directions}
We also examined whether the grafted models moved towards the reasoning-trained models in factual recall and language modelling, using TriviaQA and Wikipedia.
Grafting changed TriviaQA accuracy by $-5.0$ to $+2.8$ percentage points relative to the ungrafted models, and in the five sets where the reasoning-trained model held an advantage of 7.6 to 21.8 points the grafted model remained closer to the instruction-tuned model (Extended Data Fig.~\ref{fig:ed_knowledge}b).
Wikipedia perplexity likewise changed in both directions, by $-3.1$ to $+1.3$ (Extended Data Fig.~\ref{fig:ed_knowledge}c).
The mathematical gains thus occurred without a corresponding transfer of the donors' factual-recall advantage.

\begin{samepage}
\subsection{Reasoning gains with low interference}\label{sec:identity}

We next asked whether the grafted models retained the problems they could already solve, since a gain in overall accuracy can conceal newly incorrect answers outnumbered by newly correct ones.
We therefore tracked individual problems before and after grafting, for the region and for equally wide grafts that did not overlap it.

\end{samepage}

\subsubsection{Region grafts retained more previously solved problems than grafts elsewhere}
Across the six models and three benchmarks, region grafts retained 91.9\% of previously solved problems against 82.2\% for non-overlapping grafts (difference 9.6 percentage points; 95\% confidence interval, 7.2 to 12.1) and solved 18.1\% of previously unsolved problems against 10.0\%.

\subsubsection{Newly solved problems outnumbered newly unsolved ones in every model}
Within individual models, after pooling the three benchmarks, newly solved problems outnumbered newly unsolved ones in all six models with region grafts, whereas the net change was negative in all six with non-overlapping grafts (Table~\ref{tab:census}; Extended Data Table~\ref{tab:ed_census_detail}).
The region thus supported mathematical improvement while preserving more of the instruction-tuned models' existing correct answers than equally sized replacements elsewhere.

\begin{table}[!t]
\centering
\caption{\textbf{Higher retention and positive net change with region grafts.}
Retention is the fraction of problems solved by the no-graft host that remain solved after grafting; net is the number fixed minus the number broken.
A problem is solved when its single MATH-87 response or at least two of its four AIME responses are correct.
``Region'' denotes the graft at the MATH-87-defined span; ``None'' averages equally wide grafts with no region overlap.
Transition counts are pooled across the three mathematics benchmarks within each host and generation run, with non-overlapping graft positions averaged within host and run.
Each cell gives the mean and 95\% confidence interval across three generation runs.
Model codes follow Fig.~\ref{fig:graft}.}
\label{tab:census}
{\scriptsize
\setlength{\tabcolsep}{7pt}
\begin{tabular}{lrrrr}
\toprule
\rowcolor{gray!15}& \multicolumn{2}{c}{Retention (\%)} & \multicolumn{2}{c}{Net (fixed $-$ broken)} \\
\cmidrule(lr){2-3}\cmidrule(lr){4-5}
Host & Region & None & Region & None \\
\midrule
M3B  & $93.9\ [88.5,99.3]$  & $73.0\ [67.5,78.4]$  & $+11.3\ [1.3,21.4]$  & $-16.3\ [-23.4,-9.1]$ \\
Q7B  & $90.3\ [79.3,101.3]$ & $86.6\ [82.4,90.7]$  & $+5.0\ [-6.4,16.4]$  & $-2.3\ [-7.1,2.6]$ \\
M8B  & $93.9\ [90.7,97.0]$  & $86.5\ [84.0,89.0]$  & $+5.7\ [2.8,8.5]$    & $-4.6\ [-8.8,-0.5]$ \\
L8B  & $80.1\ [70.1,90.1]$  & $56.1\ [53.4,58.8]$  & $+5.0\ [5.0,5.0]$    & $-10.4\ [-17.4,-3.4]$ \\
Q14B & $92.0\ [91.1,92.8]$  & $84.9\ [80.9,89.0]$  & $+6.3\ [1.2,11.5]$   & $-4.3\ [-13.8,5.2]$ \\
M14B & $93.8\ [87.8,99.8]$  & $90.2\ [85.4,94.9]$  & $+5.0\ [-8.1,18.1]$  & $-2.5\ [-13.0,8.0]$ \\
\bottomrule
\end{tabular}}
\end{table}

\Needspace{8\baselineskip}
\subsection{How do reasoning-related regions develop during training?}\label{sec:trajectory}

In cortical development, the principal gradient shifts gradually over childhood and adolescence as reasoning abilities emerge~\cite{dong2021}.
This prompted us to ask whether a \reasonregion{} likewise takes shape gradually during training.
We examined OLMo~3~7B~\cite{olmo3}, whose intermediate stages of reasoning training are publicly available, and followed layers 8--14 from the base model to the end of supervised fine-tuning (Methods).
A maturity index summarized how far the displacement pattern across these layers had progressed at each stage, from zero for the base model to one at the end of training (Methods).
We asked when the pattern formed, whether its formation accompanied gains in performance, and whether grafts from the region already helped while the pattern was still forming.

\subsubsection{The regional pattern formed gradually and separated from the rest of the model}
\begin{figure}[!t]
\centering
\includegraphics[width=\textwidth]{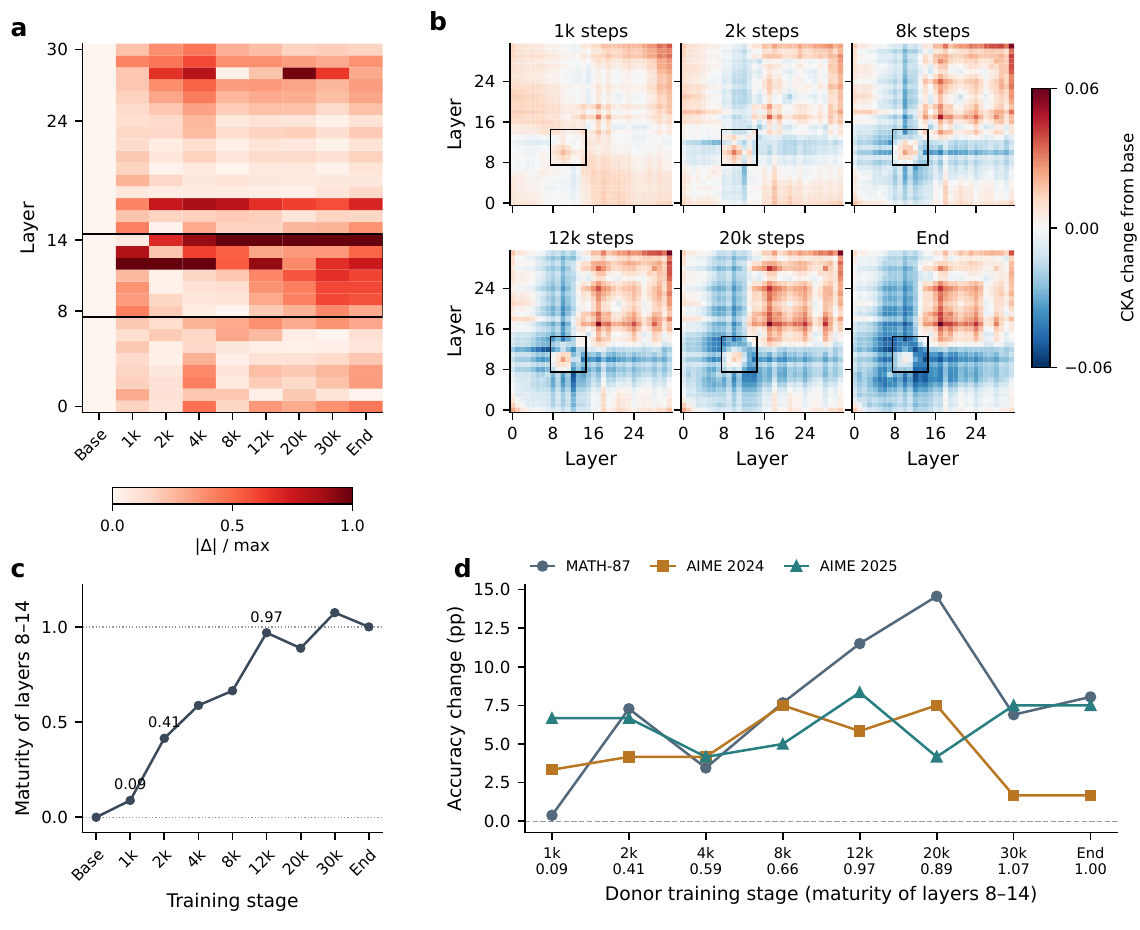}
\caption{\textbf{Regional changes during supervised reasoning training of OLMo 3 7B.}
\textbf{a}, Absolute gradient~1 displacement from the base model on MATH-87 inputs, averaged across 11 representation sites per layer and normalized to each stage's maximum (final layer excluded); the box marks layers 8--14.
\textbf{b}, Change in CKA from the base model at six stages of training, averaged over node pairs within each layer pair (self-comparisons excluded); the colour scale is clipped at $\pm0.06$ and boxes mark layers 8--14. Red, increased similarity; blue, decreased similarity.
\textbf{c}, Maturity of the layers 8--14 pattern, the projection of each stage's displacement onto the endpoint pattern; it is zero for the base model and one at the endpoint by definition and can leave this range (Methods).
\textbf{d}, Graft gains from successive stages of training. Each graft replaces layers 8--14 of the base host with the corresponding weights from one of eight donor stages (1k to 30k steps and the endpoint); positions are equally spaced in training order, and the lower line of axis labels gives the donor's maturity. Accuracy change is relative to the ungrafted base host on MATH-87 (87 problems; mean of three generation runs, whose individual gains ranged from $-4.6$ to 24.1 percentage points; host 30, 39 and 35 correct) and on AIME 2024 and AIME 2025 (four responses to each of 30 problems in one generation run; host 9 and 6 correct of 120).}
\label{fig:trajectory}
\end{figure}

Displacement appeared at layers 12--13 within the first 1{,}000 steps and at layer 14 by 2{,}000 steps, and only later at layers 8--11; the maturity index was 0.09 at 1{,}000 steps and 0.41 at 2{,}000, reached 0.97 by 12{,}000 steps and thereafter fluctuated around one (Fig.~\ref{fig:trajectory}a,c).
As training progressed, the region grew less similar to the rest of the model, the layers above it more alike and those below less so, and similarity within the region rose early and then fell below its base level (Fig.~\ref{fig:trajectory}b).

\subsubsection{Regional maturity tracked the early rise in performance}
Across the intermediate stages, the maturity index correlated with accuracy on MATH-87 ($r=0.97$), AIME~2024 ($r=0.91$) and AIME~2025 ($r=0.96$), more closely than the same index computed for any of the 13 equally wide spans that did not overlap the region, with the smallest margin on AIME~2025 (0.96 against 0.92; Extended Data Fig.~\ref{fig:ed_olmo_maturity}a,b).
After 12{,}000 steps, maturity stayed near one (Fig.~\ref{fig:trajectory}c), while AIME accuracy rose by another 13 to 18 percentage points by the endpoint (Supplementary Note ``OLMo stage accuracy and graft positions across training'').
The region's formation thus coincided with the early rise in performance, while later gains arose without a sustained increase in regional maturity.

\subsubsection{Grafts from the region helped while the pattern was still forming}
We grafted the region from each stage of training into the base model from which all stages descend (Fig.~\ref{fig:trajectory}d).
On MATH-87, the gain rose irregularly from 0.4 percentage points for the earliest donor to 11.5 at 12{,}000 steps and 14.6 at 20{,}000 steps, then fell back to 6.9 and 8.0 for the two latest donors in every generation run.
On AIME, where the base host answered only 9 and 6 of 120 responses correctly, grafts from every donor raised accuracy by about 2 to 8 percentage points with no consistent trend across stages (Methods).
The benefit on MATH-87 thus rose and fell with the donor's training stage, whereas on AIME it was present from the earliest stage.

\subsection{Do grafting gains track donor-like behaviour?}\label{sec:policy}

Returning to the six matched model sets, we asked whether hosts that received a graft had also come to behave like their donors, comparing equally wide grafts at different positions along the model.
The donor-trace score measures how far a graft's word-by-word predictions on the donor's solutions move from the host's towards the donor's.
The generated-solution score instead compares how probable the graft's own solutions are under the donor and host.
Both scores are normalized to zero for the host and one for the donor (Methods).
As before, we grouped grafts by their overlap with the region (Fig.~\ref{fig:graft}).

\subsubsection{Region grafts were more donor-like than non-overlapping grafts in five hosts}
Both scores increased from grafts with no overlap to the region graft on all three benchmarks in five hosts; M8B moved the opposite way (Extended Data Fig.~\ref{fig:ed_policy_positions}a,b).
M14B nevertheless remained farther from its donor than the ungrafted host on donor traces.
Even region grafts stayed well short of the donor: generated-solution scores reached 0.3 to 0.7 in Q7B and M3B and stayed below 0.4 in the other hosts.
In M8B, the region graft raised AIME~2024 accuracy by 7.5 percentage points while being less donor-like than grafts with no overlap.

\subsubsection{Grafts also improved on problems their donors did not solve}
On AIME, the Q7B, L8B and Q14B grafts, whose donors were distilled from the same teacher, gained almost exclusively on problems the donor solved but the host did not (Extended Data Table~\ref{tab:ed_policy_problems}).
The M3B, M8B and M14B grafts also gained on problems that neither host nor donor solved.
Correlations between problem-level gains and donor accuracy showed no consistent advantage for the own donor over the mean for the other five ($\Delta\rho$ from $-0.29$ to $+0.08$).
Grafts also rarely reproduced wrong answers that their donor gave but the host did not, with Q7B the only host where the rate rose by more than three points (Extended Data Table~\ref{tab:ed_policy_problems}).
Taken together, grafting shifted most hosts partly towards their donors' behaviour, with gains extending to problems their donors did not solve.

\subsection{Does regional grafting extend to a medical model?}\label{sec:medicalhost}

The preceding experiments concerned models trained to improve reasoning.
We extended the analysis to UltraMedical, a medical specialization of the 8B Llama instruction-tuned model~\cite{zhang2024ultramedical}, comparing its representations with those of the corresponding base model on MedQA questions~\cite{jin2021medqa} and selecting layers 13--18 from the resulting displacement profile (Methods).
We first tested whether this span transferred medical competence to the instruction-tuned parent.
We then examined whether reasoning grafts improved mathematical performance while preserving medical accuracy and existing correct answers, and how these effects depended on graft position.

\begin{figure}[!t]
\centering
\includegraphics[width=\textwidth]{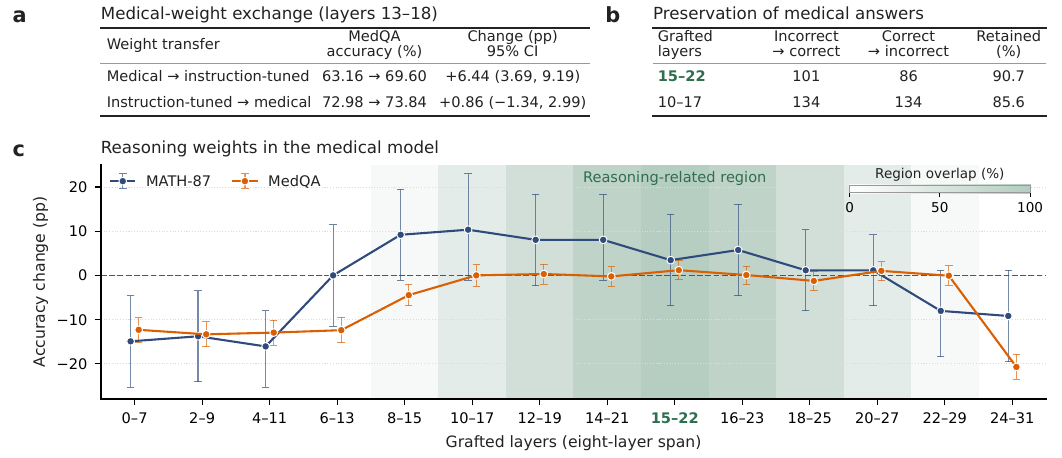}
\caption{\textbf{Regional grafting in a medical model.}
\textbf{a}, MedQA accuracy before and after exchanging weights at layers 13--18 between UltraMedical and its instruction-tuned parent, Llama-3.1-8B-Instruct.
The table lists the direction of weight transfer, recipient accuracy before and after grafting, and the change relative to the corresponding ungrafted recipient with its 95\% confidence interval (CI).
\textbf{b}, Changes in individual MedQA answers after grafting weights from DeepSeek-R1-Distill-Llama-8B at the reasoning-related region (layers 15--22, bold green row label) or at layers 10--17, the span with the largest observed MATH-87 gain.
Counts show incorrect-to-correct and correct-to-incorrect transitions relative to ungrafted UltraMedical; retention is the percentage of its 929 correct medical answers that remained correct.
\textbf{c}, Accuracy changes from ungrafted UltraMedical after eight-layer grafts from the same reasoning-trained model, evaluated on MATH-87 (navy) and MedQA (orange).
The 13 stride-two positions and the reasoning-related region previously identified using MATH-87, layers 15--22 (bold axis label), are shown.
Equally spaced categories denote zero-based, inclusive graft spans ordered by their starting layer; points are slightly offset between benchmarks for visibility.
Background green intensity indicates the percentage of the reasoning-related region included in each graft, $100|G\cap R|/|R|$, where $G$ is the grafted span and $R=$ layers 15--22; shading is constant within each category.
Circles show accuracy changes in percentage points (pp), error bars indicate 95\% CIs, and the dashed line indicates no change.
CIs in \textbf{a,c} are 95\% paired item-bootstrap intervals over the 87 mathematics and 1{,}273 medical questions (Statistics and reproducibility).}
\label{fig:medicalhost}
\end{figure}

\subsubsection{The gradient analysis located a transferable span in the medical model}
Grafting the medical model's weights at layers 13--18 into its instruction-tuned parent raised MedQA accuracy by 6.4 percentage points (Fig.~\ref{fig:medicalhost}a).
The reverse exchange, which replaced the same span in the medical model with its parent's weights, changed medical accuracy by 0.9 points.
The span thus carried competence that transferred to the parent, while the medical model's own competence did not rest on that span alone.

\subsubsection{Reasoning grafts changed medical accuracy little at intermediate positions}
We then grafted the reasoning-related region (layers 15--22) from the matched reasoning-trained model into the medical model, without further training, and evaluated MATH-87 and MedQA.
Medical accuracy stayed close to the ungrafted level, and mathematical accuracy rose by 3.4 points with an interval that included zero (Fig.~\ref{fig:medicalhost}c).
Grafts of the same width at 13 other positions depended on location: early positions reduced accuracy on both benchmarks, intermediate positions overlapping the region gave higher mathematical point estimates with little change in medical accuracy, and the final eight layers sharply reduced medical accuracy and often produced repetitive generation (Fig.~\ref{fig:medicalhost}c).

\subsubsection{The region preserved more existing medical answers than the span with the largest mathematical gain}
At layers 10--17, the position with the largest mathematical gain, equal numbers of newly correct and newly incorrect answers left overall medical accuracy unchanged, and 86\% of the previously correct medical answers remained correct, against 91\% for the region (Fig.~\ref{fig:medicalhost}b).
Within the medical model as well, the region thus traded a smaller mathematical gain for less loss of existing answers.

\FloatBarrier
\section{Discussion}\label{sec:discussion}

Cortical gradient analysis linked changes in language-model organization to locations where exchanging weights improved performance.
It described each representation through its relationships with other representations, yielding localized differences that did not track the magnitude of parameter change.
The graft experiments gave these differences functional relevance: weights from the identified spans improved mathematical reasoning on average, with outcomes depending on location, host and benchmark.
This connects a description of representational organization to an intervention on the model's computation.

The choice of inputs shaped which differences the analysis revealed.
MATH-87 elicited more consistent graft gains than AIME or non-mathematical text, suggesting that inputs within a model's capabilities may reveal functionally useful changes more clearly.
L8B's MATH-87 graft improved MATH-87 accuracy but left AIME accuracy near zero (Extended Data Table~\ref{tab:ed_other_inputs}c), showing that identifying a span need not confer reliable performance on harder problems.
Gains on held-out MATH-500 problems in five of six hosts nevertheless extended beyond the inputs used for localization.

The value of a regional graft depended on what the host retained as well as what it gained.
Higher retention alongside recovery suggests that the transferred weights could support new correct answers while working with the host's remaining layers.
The partial shift towards donor behaviour, together with gains on problems the donor failed, likewise points to an effect shaped by the receiving model.
The asymmetric medical exchange further shows that a span can carry transferable competence even when replacing it in the donor has little effect.
Preservation also mattered when choosing between graft positions: in the medical host, the region preserved more existing correct answers than the position with the largest mathematical gain.
Selecting a graft by its gain on one benchmark alone can therefore overlook disruption to competence already present in the host.

The training trajectory separates the formation of the regional pattern from subsequent gains in performance.
Grafts were already useful while the regional pattern was forming, and performance continued to improve after maturity approached its endpoint level.
The final pattern thus marks a change in organization whose formation accompanies only part of the improvement during training.
How the region interacts with the rest of the model as training proceeds remains an open question, particularly because later donor stages did not consistently yield larger graft gains.

Several limits qualify these conclusions.
Region localization depends on the reference model, inputs and selection threshold, and may be less informative when inputs exceed a model's capabilities. The selected spans need not maximize graft accuracy.
The input sets differed in content, size and difficulty, leaving the role of model competence unresolved.
Forced continuation did not consistently reproduce graft gains, but its effects varied by host and benchmark and it did not match solution lengths in every comparison.
Isolating changes specific to reasoning would require matched comparisons with other forms of post-training.
The trajectory involved one model family and one graft host, and the behavioural comparisons were descriptive.
Functional validation centred on mathematics and architecturally matched models, with a limited medical extension.

These findings motivate using representational organization to guide weight exchange while evaluating both improvement and preservation of existing answers.
Broader comparisons of selection rules across models and abilities would establish where this approach is most useful.
\backmatter


\bmhead{Author contributions}

All authors contributed to the conceptualization of the study, data analysis and manuscript writing. D.L. implemented the method and conducted the experiments. S.Yu assisted with the experiments. S.P. and H.Y.P. participated in project discussions and provided feedback. M.C. and S.Yoon jointly supervised the research. All authors reviewed and approved the final manuscript.

\bmhead{Competing interests}

The authors declare no competing interests.

\bibliography{sn-bibliography}

\section{Methods}\label{sec:methods}

\subsection{Models}

The main reasoning analysis used six architecture- and scale-matched model triplets spanning 3B to 14B parameters: three Ministral sizes, two Qwen sizes and one Llama size~\cite{ministral3,qwen2.5,llama3,deepseekr1}.
Each triplet, referred to here as a matched model set, contained a base, an instruction-tuned and a reasoning-trained model.
The Qwen and Llama reasoning models were distilled from DeepSeek-R1, whereas the Ministral reasoning models underwent supervised fine-tuning followed by reinforcement learning and preference optimization~\cite{deepseekr1,ministral3}.
The instruction-tuned model served as the host and no-graft reference because it reliably followed the evaluation prompts. Using this post-trained host was intended to reduce confounding from instruction following and response formatting relative to a base-model host.
The base--reasoning comparison defined the gradient-displacement profile and the \reasonregion{}; we then assessed its functional relevance by replacing the corresponding layer weights in the instruction-tuned model with those of the reasoning-trained model. Matching within each model set ensured identical layer structure and parameter dimensions. Exact model identifiers, layer counts, representation-node counts, and architecture and projection dimensions are reported in Supplementary Table~1.

The medical analysis additionally used Llama-3.1-8B-UltraMedical~\cite{zhang2024ultramedical}, a medical specialization of Llama-3.1-8B-Instruct.
Its base-model reference was Llama-3.1-8B, and reasoning weights came from DeepSeek-R1-Distill-Llama-8B.
These models share 32 decoder layers with matching parameter dimensions.

The training-trajectory analysis used OLMo~3~7B~\cite{olmo3}: the pretrained model and the published intermediate stages of its supervised fine-tuning for reasoning, whose training continues with preference optimization and reinforcement learning after the stages examined here (Training-trajectory analysis).

\subsection{Datasets}

Mathematical reasoning was evaluated on three benchmarks: MATH-87, a fixed 87-problem subset of MATH-500~\cite{hendrycks2021math,lightman2023} restricted to statements of 200--300 characters to reduce variation in input length; AIME~2024~\cite{aime2024} (30 problems); and AIME~2025~\cite{aime2025} (30 problems from AIME~2025-I and II).
Factual recall was evaluated on 500 TriviaQA questions~\cite{joshi2017}, and language modelling on 100 Wikipedia passages of 200--300 words.

For gradient comparisons using non-mathematical inputs, we used 100 English Wikipedia texts of 30--50 words and 100 random word sequences of the same length range.
Wikipedia texts were sampled with seed 17 from the November 2023 English Wikipedia release.
Random sequences were generated with seed 42 by choosing a length between 30 and 50 words and sampling English words with replacement from a general English word list.
Each input set was shared across the compared models.

Medical performance was evaluated on the complete 1{,}273-question, four-option MedQA-USMLE test set~\cite{jin2021medqa}.
For gradient analysis, we selected 87 questions from this test set whose question and option text contained 176--212 tokens under the Llama-3.1-8B-Instruct tokenizer, excluding special tokens.
This range covered the 40th--60th percentiles of test-question length.
Sampling used seed 17 and was stratified by the correct option.
The fixed dataset version and selected question identifiers are included in the analysis source data.

\subsection{Activation sampling}

We sampled activations at 11 predefined sites in each transformer layer, listed in forward-computation order: in self-attention, the normalized attention input, the query, key and value projections, the concatenated attention context before the output projection, and the output projection; in the feed-forward network (FFN), the normalized FFN input, the gate and up projections, the gated intermediate before the down projection, and the down projection. A representation node comprised one layer--site pair, yielding $U=11L$ nodes for a model with $L$ layers.

At each node $u$, activation vectors from the same valid, non-padding token positions across all input texts were stacked into a matrix $\mathbf{H}_u\in\mathbb{R}^{N\times d_u}$. The observation count $N$ and row order were shared across nodes, whereas the feature dimension $d_u$ could vary by site and model architecture.

\subsection{Representational similarity}

Representational similarity between all pairs of representation nodes was measured with linear centered kernel alignment (CKA)~\cite{kornblith2019}, which is invariant to orthogonal transformation and isotropic scaling. For each node $u$, we formed the linear Gram matrix $\mathbf{G}_u=\mathbf{H}_u\mathbf{H}_u^{\top}$ and centred it as $\mathbf{K}_u=\mathbf{C}\mathbf{G}_u\mathbf{C}$, where $\mathbf{C}=\mathbf{I}_N-N^{-1}\mathbf{1}\mathbf{1}^{\top}$. For nodes $u$ and $v$, we computed
\begin{equation}
\operatorname{CKA}(\mathbf{H}_u,\mathbf{H}_v)
=\frac{\operatorname{HSIC}(\mathbf{K}_u,\mathbf{K}_v)}
{\sqrt{\operatorname{HSIC}(\mathbf{K}_u,\mathbf{K}_u)\,
\operatorname{HSIC}(\mathbf{K}_v,\mathbf{K}_v)}}.
\end{equation}
Here $\operatorname{HSIC}(\mathbf{K}_1,\mathbf{K}_2)=(N-1)^{-2}\operatorname{tr}(\mathbf{K}_1\mathbf{K}_2)$. Pairwise scores formed the representational-similarity matrix $\mathbf{S}_{\mathrm{CKA}}\in\mathbb{R}^{U\times U}$. Because all Gram matrices contain the same $N$ observations, CKA can compare representation nodes with different feature dimensions.

\subsection{Diffusion-map embedding}

Following methods developed for brain connectivity~\cite{margulies2016}, we computed a diffusion-map embedding~\cite{coifman2006} from $\mathbf{S}_{\mathrm{CKA}}$.
We took absolute values, set the diagonal to one, retained the largest 10\% of entries in each row and computed cosine similarities between the sparsified rows to obtain an affinity matrix $\mathbf{W}$.
Row normalization gave the Markov transition matrix $\mathbf{P}$, whose right eigenvectors provided the gradient coordinates:
\begin{equation}
P_{uv}=\frac{W_{uv}}{\sum_{v'=1}^{U}W_{uv'}},
\qquad
\mathbf{P}\boldsymbol{\psi}_k=\lambda_k\boldsymbol{\psi}_k.
\label{eq:diffusion_coordinates}
\end{equation}
Eigenvalues were ordered from largest to smallest; the leading eigenvector, with eigenvalue one, is the trivial component and was discarded, and the remaining eigenvectors were indexed from one.
The model analyses used these non-trivial eigenvectors without eigenvalue scaling; $\boldsymbol{\psi}_1$ defines gradient~1.

\subsection{Gradient alignment and layer-wise displacement}

Within each matched model set, we aligned the instruction-tuned and reasoning-trained models to the base model using their first three non-trivial eigenvectors.
For model $m$, let $\boldsymbol{\Psi}^{(m)}=[\boldsymbol{\psi}_1^{(m)},\boldsymbol{\psi}_2^{(m)},\boldsymbol{\psi}_3^{(m)}]\in\mathbb{R}^{U\times3}$ and let $\mathbf{X}^{(m)}$ be its column-centred form.
The orthogonal Procrustes rotation~\cite{schonemann1966} was fitted across all representation nodes:
\begin{equation}
\mathbf{Q}_m^{\star}
=\underset{\mathbf{Q}^{\top}\mathbf{Q}=\mathbf{I}_3}{\arg\min}
\left\|\mathbf{X}^{(m)}\mathbf{Q}-\mathbf{X}^{(\mathrm{base})}\right\|_{\mathrm{F}}^2,
\label{eq:gradient_alignment}
\end{equation}
where $\|\cdot\|_{\mathrm{F}}$ is the Frobenius norm.
We rotated $\mathbf{X}^{(m)}$ by $\mathbf{Q}_m^{\star}$, then translated the coordinates to match the base-model mean across the 11 sites in the first layer.
Each component was sign-matched to the base model by its Pearson correlation across nodes.
Finally, gradient~1 was given a common sign across the three models, chosen from the mean slope of the reasoning-trained profile after omitting the final 10\% of layers.

Let $\widetilde{\boldsymbol{\Psi}}^{(m)}$ denote these aligned and oriented coordinates.
We averaged the first component across the 11 sites within each layer:
\begin{equation}
g_\ell^{(m)}=\frac{1}{11}\sum_{j=1}^{11}
\widetilde{\Psi}^{(m)}_{(\ell,j),1},
\label{eq:layer_gradient}
\end{equation}
where $(\ell,j)$ identifies the representation node at site $j$ in layer $\ell$.
Base--reasoning displacement was then calculated using equation~(\ref{eq:gradient_displacement}).
The base model provides a pretrained reference; using the instruction-tuned model as reference measures differences between two post-trained models. Supplementary Note ``Reference and threshold dependence'' and Supplementary Table~2 compare these selections without attributing the difference exclusively to reasoning training.
The final transformer layer was excluded from the six-set displacement analyses.
Unsmoothed displacement values were used for Fig.~\ref{fig:detection}, the summaries in Fig.~\ref{fig:detection}g and the parameter-distance comparisons.

\subsection{Region definition}

For span selection, we smoothed $d_\ell=|\Delta_\ell|$ with a three-layer moving average to obtain $\bar d_\ell$.
In the six-set reasoning analysis, layers qualified when
\begin{equation}
\bar d_\ell\geq f\max_{r\in\mathcal{L}}\bar d_r,
\label{eq:region_threshold}
\end{equation}
where $\mathcal{L}=\{0,\ldots,L-2\}$ contains the analysed layers, excluding the final transformer layer.
We selected the longest contiguous run of qualifying layers, using $f=0.70$ except for L8B.
For L8B, the broad MATH-87 displacement profile yielded a span covering 15 of 32 transformer layers at this threshold.
To limit the extent of donor-weight replacement, we used $f=0.85$ for L8B, selecting layers 15--22 (eight of 32 layers).
A minimum width of 15\% of the analysed layers, rounded to the nearest integer, was enforced by placing a window of that width centred on the smoothed peak and shifting it to stay within the analysed layers.
The final MATH-87 spans are listed in Fig.~\ref{fig:detection}g.
Supplementary Table~3 reports how the selected spans change across seven thresholds, with all other selection rules fixed; these calculations do not evaluate graft performance at the resulting spans.
Each MATH-87 span was held fixed across the graft evaluations on MATH-87, AIME~2024 and AIME~2025 and is referred to as the \reasonregion{} following its graft evaluation.

\subsection{Layer-wise parameter-distance control}

To determine whether gradient displacement was a proxy for the overall magnitude of parameter change, we compared architecturally matched models at the parameter level. For base model $B$ and target model $T$, let $P_\ell$ contain all parameters within transformer layer $\ell$. We calculated
\begin{equation}
D_\ell(B,T)=
\frac{\sqrt{\sum_{p\in P_\ell}\lVert\theta^T_p-\theta^B_p\rVert_F^2}}
{\sqrt{\sum_{p\in P_\ell}\lVert\theta^B_p\rVert_F^2}}.
\end{equation}
This relative Frobenius distance gives equal weight to each scalar parameter. For each model set, we compared the base--instruction and base--reasoning parameter-distance profiles with the corresponding raw, unsmoothed gradient-displacement profiles obtained on MATH-87 after excluding the final transformer layer. Agreement between each pair of profiles was summarized with the Spearman rank correlation.
For the base--reasoning transition, we also compared the mean value inside the region with that outside it for each profile.

\subsection{Region grafting}

A graft copied weights from the reasoning-trained model into the corresponding layers of the architecturally matched instruction-tuned model.
It replaced all parameters of the selected transformer layers, including attention and feed-forward computations, while retaining all remaining parameters.
No further training was applied.

\subsection{Graft positions across layers}

To map how the graft effect varied with location, we retained each model's MATH-87 region width and advanced the graft's starting layer by two at a time.
Starting layers shared the same odd--even alignment as the MATH-87-defined region.
The positional, overlap and transition analyses included 76 positions from this sweep, including the six region grafts, plus nine comparison grafts from benchmark-specific selection that lay on the same grid, giving 85 positions per benchmark (Fig.~\ref{fig:graft} and Extended Data Fig.~\ref{fig:ed_stride_profiles}).
Comparison grafts off the grid entered only the donor-likeness analyses.
For each grafted layer span $G$ and reasoning-related region $R$, overlap was $|G\cap R|/|R|$, with bins defined as none ($0$), low ($(0,0.5)$), high ($[0.5,1)$) and full ($1$).

\subsection{Alternative layer-selection profiles}

We used the existing position sweep to compare region grafts with equally wide grafts selected from two base--reasoning change profiles.
The first was relative Frobenius distance, defined above.
The second measured CKA similarity-profile change: for each representation node, we took the Euclidean distance between corresponding rows of the base and reasoning-trained models' CKA matrices, then averaged these distances across the 11 nodes in each layer.
Each row contained similarities to all $U$ nodes, including itself, from the matrices used for gradient analysis before sparsification and embedding.
For both profiles, we excluded the final layer and applied a three-layer moving average, repeating the edge value at each boundary.

For each host, graft width $w$ was fixed to its reference width in the positional sweep.
Given the smoothed peak at layer $p$, we set the starting layer to $p-\lfloor(w-1)/2\rfloor$, restricted it to $[0,L-w]$, and moved it to the nearest starting layer on the sweep's two-layer grid; ties were resolved towards the lower layer index.
Graft spans could therefore include the final transformer layer.
Positions were selected from these profiles without using graft accuracy and held fixed across the three benchmarks (Extended Data Fig.~\ref{fig:ed_stride_profiles}d).
We compared the corresponding three-run mean accuracies descriptively; the comparison held width fixed and did not evaluate spans obtained by thresholding the alternative profiles. Supplementary Note ``Alternative span selection'' and Supplementary Table~4 additionally compare a central-window baseline; in M14B this window coincides with the region.

\subsection{Benchmark-specific region selection}

For the input comparison in Extended Data Table~\ref{tab:ed_other_inputs}a, we selected a span separately from each benchmark's base--reasoning displacement profile using the region definition above.
For each matched model set and input benchmark, we enumerated contiguous spans of the same width that did not overlap the selected span, then sampled three starting positions without replacement using seed 0.
Weights from the same reasoning-trained model were used for the selected span and all three comparison grafts.
Each span was identified and evaluated using the same benchmark, and its graft accuracy was compared with the mean of these three comparison grafts.
The comparison spans could therefore change when the input benchmark changed.

\subsection{Regions from non-reasoning inputs}

We used base--reasoning displacement profiles obtained on Wikipedia text and random word sequences to identify alternative layer spans with the same contiguous-span procedure.
These spans could differ in both position and width from the MATH-87 region.
For each model pair, all alternative grafts copied weights from the same reasoning-trained model into the corresponding instruction-tuned model.
These grafts were evaluated in three generation runs in five models and in one run in L8B, whose AIME accuracies were near zero, with the sampling settings described under Generation and evaluation.
Ungrafted and MATH-87 region references came from the main mathematical evaluations; alternative grafts and references were generated separately.
The evaluated spans are listed in Extended Data Table~\ref{tab:ed_other_inputs}b.

\subsection{Generation and evaluation}

Each mathematics problem was paired with an instruction to reason step by step and place the final answer in a boxed expression, then formatted with the model's chat template. The repeat analyses used three generation runs with temperature 0.7 and top-$p=0.9$ (Statistics and reproducibility). The main graft analysis and the evaluation across graft positions used one response per MATH-87 problem and run; AIME~2024 and AIME~2025 used four responses per problem and run. Generated-length analyses used these same sampled responses, with correct responses examined separately. For these repeat analyses, the requested maximum was 16{,}384 new tokens, except for Q7B, whose shorter context length limited responses to 4{,}096 tokens.

In the main three-run analyses, correctness was assigned by comparing a deterministic answer-extraction verdict with a GPT-4.1 mini verdict for each response. Agreement determined the final label, whereas disagreements were passed to GPT-5 for adjudication. All sampled responses contributed to accuracy.

The deterministic scorer selected the last complete boxed answer, or searched for a stated answer near the end of the response when no complete box was found.
It compared normalized answers using numeric or symbolic equivalence and a normalized-string fallback; failure to extract an answer counted as incorrect at this tier, not necessarily in the final adjudication.
Both LLM judges received the problem, gold answer and model response.
The first judge was prompted to require a boxed answer, whereas the adjudicator could accept an equivalent answer stated in prose.

Supplementary Note ``Evaluation units and prompt structure'' and Supplementary Table~5 give the instruction wording, placement and response denominators for the distinct mathematical evaluations.

The benchmark-specific region comparison in Extended Data Table~\ref{tab:ed_other_inputs}a used four responses per problem and condition: 348 on MATH-87 and 120 on each AIME benchmark.
These estimates are reported descriptively, separately from the three-run uncertainty analysis.

Within each generation run, accuracy was the fraction of correct sampled responses. For the benchmark-specific region selection in Extended Data Table~\ref{tab:ed_other_inputs}a, $\Delta_c=a_{\text{region}}-\bar{a}_{\text{comparison}}$, where $\bar{a}_{\text{comparison}}$ is the arithmetic mean across the three size-matched comparison grafts. For transition analyses, a MATH-87 problem was solved when its single response was correct, whereas an AIME problem was solved when at least two of four responses were correct. Retention was the fraction of problems solved by the no-graft host that remained solved after grafting, recovery was the fraction of unsolved problems that became solved, and net change was the number fixed minus the number broken. For Table~\ref{tab:census}, transition counts were pooled across the three benchmarks within host and run, and all graft positions with no region overlap were averaged within host and run. Overall retention and recovery additionally pooled these counts across hosts within each run before calculating rates and paired differences; headline estimates and intervals then summarized the three runs. Extended Data Table~\ref{tab:ed_census_detail} reports the same measures separately by benchmark and overlap class.

\subsection{Forced generation length and held-out problems}

To test whether longer solutions alone reproduced the graft gains, each ungrafted instruction-tuned model was made to continue generating: whenever a response ended before a set number of generated tokens, the word ``Wait'' was appended and generation resumed, up to eight times on MATH-87 and sixteen on AIME, under the generation limit of 16{,}384 tokens that applied to every condition.
Responses could remain below the target length when generation stopped for repetition or the allowed resumptions were exhausted.
Two target lengths were set per host and benchmark using earlier region-graft evaluations and held fixed for this comparison (Extended Data Table~\ref{tab:ed_forcing}).
MATH-87 targets were chosen to cover typical region-graft lengths; on AIME, the lower target was the median response length, rounded to 100 tokens, or the median of responses ending before the generation limit when the overall median reached it. The upper AIME target was 1.5 times the lower, capped at 12{,}000.
Q7B was not forced: its continued generations consisted of unrelated text, and its 4{,}096-token context left no room for longer solutions.
The forced conditions and their ungrafted and region-graft references were generated together in three generation runs, with the instruction and problem given in a single user turn, and are compared only with one another.
Responses that reached the generation limit were scored as generated.

For the held-out evaluation, the 413 MATH-500 problems outside MATH-87 were answered once, in the first generation run, by each ungrafted model, its region graft and its reasoning-trained model, with the same single-turn format and scoring; the 87 selection problems were answered in the same runs for comparison.

\subsection{Factual recall and language modelling}

For each of the six matched model sets, we evaluated the ungrafted instruction-tuned model, its MATH-87 region graft and the reasoning-trained model on the same TriviaQA questions and Wikipedia passages (Extended Data Fig.~\ref{fig:ed_knowledge}b,c).
We sampled 500 questions without replacement from the no-context TriviaQA validation split using seed 42.
Prompts requested only the answer, without an explanation, and used each model's chat template.
Each question received one greedily decoded response with a maximum of 400 new tokens.
A response was scored as correct if it contained any of the dataset's answer aliases or normalized aliases, using case-insensitive substring matching.
Accuracy was the fraction of the 500 responses scored as correct.

For language modelling, we used the first 100 passages in our fixed Wikipedia sample of 200--300-word texts.
Passages were tokenized separately for each model, truncated to at most 2{,}048 tokens and evaluated without a chat prompt.
For each passage, we computed the mean next-token negative log-likelihood.
Reported perplexity was the exponential of the arithmetic mean of these passage-level losses, giving each passage equal weight.
These evaluations used one fixed set of questions and passages per model condition and were reported descriptively, separately from the three-run mathematics analyses.

\subsection{Training-trajectory analysis}
The trajectory analysis used the 32-layer pretrained model Olmo-3-1025-7B and the published checkpoints of Olmo-3-7B-Think-SFT~\cite{olmo3} at steps 1{,}000, 2{,}000, 4{,}000, 8{,}000, 12{,}000, 20{,}000 and 30{,}000 and at the end of supervised fine-tuning, referred to as stages.
We used MATH-87 inputs, centred linear CKA and diffusion-map embedding as described above.
OLMo uses post-normalized transformer blocks: the attention- and feed-forward-input sites sampled the residual stream, and their output sites preceded post-normalization; the remaining sites followed the definitions above, yielding 352 representation nodes.
Procrustes alignment used all representation nodes for translation as well as rotation, and displacement was averaged over the 11 sites per layer.
Signed CKA maps show stage-minus-base differences averaged over the node pairs in each layer pair, excluding a node's similarity with itself.
Let $s_{\ell,k}$ denote the signed gradient~1 displacement from the base model at layer $\ell$ and stage $k$, and let $K$ denote the final stage.
For a fixed layer span $W$, regional maturity was
\begin{equation}
m_W(k)=\frac{\sum_{\ell\in W}s_{\ell,k}s_{\ell,K}}{\sum_{\ell\in W}s_{\ell,K}^{2}}.
\label{eq:regional_maturity}
\end{equation}
The index is zero for the base model and one at the endpoint by construction, and can exceed one.
We held layers 8--14 fixed throughout the trajectory and stage-graft analyses and calculated the same index for each of the 13 seven-layer spans that did not overlap layers 8--14, advancing their starting layer by one (Extended Data Fig.~\ref{fig:ed_olmo_maturity}b). These maturity comparisons retain all 32 layers, including the span 25--31; the displacement display in Fig.~\ref{fig:trajectory}a excludes the final layer.
Ungrafted stage evaluations used a 30{,}000-token generation limit, temperature 0.7 and top-$p$ 0.9.
On MATH-87, each evaluation generated one response per problem; reported counts averaged two generation runs for the base model, the 1{,}000-, 2{,}000- and 4{,}000-step stages and the endpoint, and used one run for the remaining stages.
Each AIME evaluation generated four responses per problem in one run, giving 120 responses per year.
Scoring used deterministic final-answer extraction and counted responses that reached the token limit or met the repetition-stop criterion as incorrect.
Stage accuracy and graft accuracy were evaluated under their respective generation limits and were not pooled.

For stage grafts, we copied all parameters at layers 8--14 from each of the eight donor stages into the pretrained model, which served as the host because every stage descends from it, retaining the host's remaining weights.
Generation used plain-text prompts, temperature 0.7, top-$p$ 0.9 and a 16{,}384-token response limit; the repetition-stop criterion was applied after generation, and correctness used the deterministic scoring described above.
MATH-87 used one response per problem in each of three generation runs; AIME used four responses per problem in one run.
Grafts and their ungrafted references shared the same problem-specific random seeds.
Gains were differences in correct-response counts from the ungrafted host, averaged over runs for MATH-87 and reported separately for the two AIME years, and are expressed as percentage points of 87 or 120 responses.
Supplementary Note ``OLMo stage accuracy and graft positions across training'' and Supplementary Table~6 extend the MATH-87 comparison to all 13 seven-layer spans starting at layers 0, 2, \ldots, 24, using the same eight donor stages, three generation runs and scoring; Supplementary Table~7 gives the answer transitions for endpoint-donor grafts. The 8--14 observations are shared with Fig.~\ref{fig:trajectory}d; this positional grid is distinct from the non-overlapping structural-maturity spans above.

\subsection{Donor similarity on generated solutions}

The behavioural analyses used the six instruction-tuned hosts ($B$), their reasoning-trained donors ($D$) and grafted models ($G$).
For each model set, we scored the MATH-87 region graft, every two-layer-step position and the three comparison grafts of the benchmark-specific selection, giving 12, 13, 22, 11, 15 and 21 distinct positions for Q7B, L8B, Q14B, M3B, M8B and M14B, respectively.
All grafts within a model set had the same width.
Accuracy changes came from the positional evaluation and averaged three generation runs, with one response per MATH-87 problem and four per AIME problem per run.
Q7B responses in that evaluation had a 4{,}096-token generation limit, and other models had a 16{,}384-token limit.

For the donor-trace comparison, each donor generated one solution per problem in one generation run with temperature 0.7 and top-$p$ 0.9.
We teacher-forced the same response tokens through $B$, $D$ and $G$, using each model's own prompt template and its full, untempered next-token distribution.
At response position $t$ of problem $i$, we calculated Kullback--Leibler (KL) divergence from $D$ to $B$ and from $D$ to $G$.
The fraction of host--donor divergence closed by a graft was
\begin{equation}
R_G=1-\frac{\sum_{i,t}\operatorname{KL}(p_{D,i,t}\Vert p_{G,i,t})}
{\sum_{i,t}\operatorname{KL}(p_{D,i,t}\Vert p_{B,i,t})}.
\label{eq:donor_trace_similarity}
\end{equation}
Thus $R_G=0$ means equal total divergence from the donor for the graft and host, $R_G=1$ means agreement with the donor on the scored prefixes, and negative values indicate greater divergence than the host.
We refer to $R_G$ as the donor-trace score.
Chat-template tokens were excluded from the reported scores.
For MATH-87, sequences exceeding 12{,}288 tokens including the prompt were omitted; the retained problem counts are given in the legend of Extended Data Fig.~\ref{fig:ed_policy_positions}.
For AIME, Q7B sequences were truncated at 4{,}096 tokens, whereas other models used a 20{,}480-token scoring limit with over-limit sequences omitted.

For the generated-response comparison, we used the first response of the first generation run from each graft and host in the positional evaluation, and the donor traces as donor reference responses.
For the tokens generated by a model $X$, we calculated
\begin{equation}
S_X=\frac{1}{N_X}\sum_{i,t}\left[\log p_D(y^X_{i,t}\mid y^X_{i,<t})-\log p_B(y^X_{i,t}\mid y^X_{i,<t})\right],
\qquad \operatorname{pos}_G=\frac{S_G-S_B}{S_D-S_B},
\label{eq:generated_response_similarity}
\end{equation}
where $N_X$ is the total number of scored response tokens, and conditioning on the question and each scoring model's prompt is implicit.
We refer to $\operatorname{pos}_G$ as the generated-solution score.
Chat-template tokens were excluded.
Q7B sequences were truncated to a 4{,}096-token scoring context; other models used 20{,}480 tokens.

Overlap-class values averaged the positions in each class, with the region graft providing full overlap. Supplementary Note ``Interpreting donor-likeness scores'' and Supplementary Table~8 separate the scores' conditioning, sampling and numerical values.
For the AIME problem-level comparison, we classed each problem by whether the host's modal answer over its 12 responses and the donor's modal answer over nine responses were correct, and reported the region graft's mean change in the fraction of correct responses over the problems the host did not solve, separately for problems the donor solved and failed.
Final-answer analyses used deterministic extraction, with 12 responses per graft and host (three generation runs, four responses per problem).
To identify a donor's distinctive wrong answers, we used its modal answer over nine responses: the donor trace and eight previously collected donor responses.
Eligible problems had a parseable, incorrect donor mode differing from the host mode; the shared-wrong rate was the fraction of graft responses matching that donor answer.
For donor specificity, we correlated problem-level gains with each donor's accuracy over the eight previously collected responses, restricting to problems the host had not answered correctly in all 12 responses, and subtracted the mean correlation with the other five donors from that with the own donor.
The previously collected donor responses used a different prompt placement from the donor traces.

\subsection{Medical-model analysis and grafting}
We compared Llama-3.1-8B and UltraMedical using the 87-question MedQA subset and the activation sampling, CKA, diffusion-map embedding and alignment procedures described above, and selected the longest contiguous span of elevated smoothed displacement using $f=0.70$, yielding layers 13--18.
Weights from this span were exchanged in both directions between UltraMedical and Llama-3.1-8B-Instruct, and each graft was compared with its ungrafted recipient on the full MedQA test set.

To examine reasoning weights in the medical model, we grafted layers 15--22 from DeepSeek-R1-Distill-Llama-8B into UltraMedical; this eight-layer reasoning-related region was identified using MATH-87 and retained without re-estimating its boundaries on medical inputs.
We also evaluated eight-layer grafts starting at layers 0, 2, \ldots, 24, giving 14 positions including the region graft, on MATH-87 and the full MedQA test set.
Each graft replaced all parameters within the selected layers while retaining the recipient's remaining parameters, without further training; layer indices are zero-based and inclusive.

Generation used temperature 0.7, top-$p=0.9$ and one response per question in one generation run.
Medical prompts presented the question and four options and requested a final answer letter A--D; mathematical prompts followed the instructions described under Generation and evaluation.
Grafted models and the ungrafted instruction-tuned reference used a 4{,}096-new-token limit; ungrafted UltraMedical responses, generated with a 16{,}384-token limit, were truncated to 4{,}096 tokens before scoring.
Generation also stopped when the last 256 generated tokens contained three consecutive repetitions of a 5--50-token sequence.

Mathematical answers were assessed using the deterministic and LLM-based judgments described above.
Medical answers were extracted deterministically from the final lines of each response, accepting a stated letter, a boxed letter or the text of a single option, and compared with the correct option; responses without a unique choice were counted as incorrect, and all responses remained in the accuracy denominator, including those stopped at the token limit or for repetition.
Medical retention was the fraction of the ungrafted recipient's correct answers that remained correct after grafting, and the comparison in Fig.~\ref{fig:medicalhost}b used the reasoning-related region and layers 10--17, selected from the position sweep as the span with the largest observed MATH-87 gain.

\subsection{Statistics and reproducibility}
Generation runs used the random seeds 101, 202 and 303 in that order; where fewer runs were used, they were the first of these.
For the main six-host graft and positional analyses, the generation run was the repeat unit for uncertainty estimates, and the generation runs named in the main text and captions are these repeats.
Graft and no-graft conditions were paired within host and run; benchmark-level summaries first averaged the six hosts equally within each run.
For overlap-bin summaries, graft positions were first averaged within host, overlap class and run; the region graft supplied the full-overlap condition.
Points report the mean of the three runs, and intervals are two-sided 95\% Student-$t$ confidence intervals across the three run-level estimates ($df=2$), without clipping to metric bounds.
Individual run estimates are shown in the aggregate accuracy and response-length summaries (Fig.~\ref{fig:graft}a and Extended Data Fig.~\ref{fig:ed_knowledge}a).
For the comparison with other positions (Fig.~\ref{fig:graft}b), the region graft was compared with the mean of all other graft positions sharing its starting-layer parity within host and run.
For the length analysis, each graft's median was divided by the corresponding no-graft median within model and generation run.
Separate ratios used all responses or only correct responses in both the grafted and ungrafted conditions; the correct-response subsets could therefore contain different problems.
Medians and ratios were first averaged across graft positions within model, overlap class and run.
The ratio summaries then weighted models equally within each run before summarizing the three runs.
Correct-response ratios were undefined when either condition had no correct response; undefined values were omitted at each averaging step, so the contributing positions and models could vary.
These length summaries are descriptive because generation length is measured after grafting.
Individual problems, graft positions and responses were not treated as independent replicates.
In the trajectory analysis, Pearson correlations between maturity and accuracy used the seven intermediate stages, excluding the endpoint because its maturity is fixed at one; stages and overlapping comparison spans are not independent replicates, and these associations are descriptive.
The donor-likeness, problem-level and answer comparisons are descriptive and carry no uncertainty intervals.
In the medical analysis, changes in correctness were paired by question, and 95\% confidence intervals for accuracy differences were the 2.5th and 97.5th percentiles of 10{,}000 paired item-bootstrap resamples within each benchmark (bootstrap seed 0); these pointwise intervals quantify item-sampling uncertainty in a single generation run and are not adjusted for the position search.

\Needspace{12\baselineskip}
\section{Data availability}

The source models and evaluation benchmarks used in this study are publicly available.
Pre-computed similarity matrices will be released upon publication.
Analysis source data, including fixed input selections, model and dataset versions, response-level correctness records and values underlying the figures and tables, are available from the corresponding authors upon reasonable request, together with the grafted models.

\section{Code availability}

Code for activation sampling, CKA computation, diffusion-map embedding, region selection, weight grafting, response generation and scoring, statistical analysis and figure reproduction will be provided to editors and reviewers during peer review and made publicly available upon publication.

\clearpage
\section{Extended Data}
\setcounter{table}{0}
\renewcommand{\tablename}{Extended Data Table}
\renewcommand{\thetable}{\arabic{table}}
\providecommand{\theHtable}{}
\renewcommand{\theHtable}{ED.\arabic{table}}

\vspace*{\fill}
\begin{center}
\captionof{table}{\textbf{Grafts at spans identified from other inputs.}
For each model pair, all grafts copied weights from the same reasoning-trained model into the matched instruction-tuned model (host), without further training.
\textbf{a}, A span was identified separately from each benchmark's base--reasoning displacement profile and evaluated on the same benchmark, using four responses per problem in one generation run.
Within each evaluation, $\Delta_c$ is the selected-span graft's accuracy minus the mean accuracy of three randomly drawn grafts of the same width outside that span, in percentage points; the last row counts models with $\Delta_c>0$.
$^{\ddagger}$The ungrafted reference accuracy of L8B was 6/120 on AIME~2024 and 0/120 on AIME~2025.
\textbf{b}, Layer spans (zero-based, inclusive) identified from base--reasoning displacement profiles on Wikipedia text and random word sequences. Their positions and widths could differ from the MATH-87 region; parentheses give the fraction of that region covered by each span.
\textbf{c}, Accuracy after grafting at the Wikipedia and random-word spans in \textbf{b}. The Region column uses the MATH-87-defined span, held fixed across all three benchmarks; No graft denotes the ungrafted host. Values are means and 95\% confidence intervals across three generation runs (MATH-87, one response per problem and run; AIME, four). $^{\dagger}$Alternative grafts in L8B were evaluated in one generation run only; its ungrafted and region references use all three runs.
Model codes follow Fig.~\ref{fig:graft}.}
\label{tab:ed_other_inputs}
\par\vspace{6pt}
\begin{minipage}[t]{0.49\textwidth}\centering
{\scriptsize\setlength{\tabcolsep}{3.5pt}
\begin{tabular}{lrrr}
\multicolumn{4}{l}{\textbf{a}\quad\textbf{Grafts at spans selected per benchmark ($\Delta_c$, pp)}}\\[3pt]
\toprule
\rowcolor{gray!15}Host & MATH-87 & AIME~2024 & AIME~2025 \\
\midrule
M3B  & $+19.06$ & $+10.28$          & $-3.61$ \\
Q7B  & $+5.56$  & $+1.94$           & $+1.39$ \\
M8B  & $+7.95$  & $+3.61$           & $-8.89$ \\
L8B  & $+17.53$ & $0.00^{\ddagger}$  & $-0.83^{\ddagger}$ \\
Q14B & $+6.61$  & $-1.94$           & $-0.83$ \\
M14B & $-0.38$  & $-6.11$           & $-3.33$ \\
\midrule
Models with $\Delta_c>0$ & 5/6 & 3/6 & 1/6 \\
\bottomrule
\end{tabular}}
\end{minipage}\hfill
\begin{minipage}[t]{0.49\textwidth}\centering
{\scriptsize\setlength{\tabcolsep}{4pt}
\begin{tabular}{llll}
\multicolumn{4}{l}{\textbf{b}\quad\textbf{Spans identified from non-reasoning inputs}}\\[3pt]
\toprule
\rowcolor{gray!15}Host & Region & Wikipedia & Random words \\
\midrule
M3B & 16--24 & 10--15 (0) & 8--12 (0) \\
Q7B & 12--18 & 10--20 (1.00) & 21--26 (0) \\
M8B & 25--31 & 28--32 (0.57) & 8--13 (0) \\
L8B & 15--22 & 7--12 (0) & 1--5 (0) \\
Q14B & 25--31 & 29--36 (0.43) & 27--33 (0.71) \\
M14B & 16--21 & 26--35 (0) & 8--13 (0) \\
\bottomrule
\end{tabular}}
\end{minipage}\\[8pt]
{\scriptsize\setlength{\tabcolsep}{5pt}
\begin{tabular}{llrrrr}
\multicolumn{6}{l}{\textbf{c}\quad\textbf{Accuracy after grafting (\%)}}\\[3pt]
\toprule
\rowcolor{gray!15}Host & Benchmark & No graft & Region & Wikipedia & Random words \\
\midrule
M3B & MATH-87 & $78.9\ [71.7,86.1]$ & $82.8\ [77.8,87.7]$ & $78.5\ [74.2,82.9]$ & $70.9\ [66.5,75.2]$ \\
M3B & AIME 2024 & $26.7\ [21.2,32.1]$ & $35.8\ [32.2,39.4]$ & $26.4\ [21.2,31.6]$ & $17.8\ [13.5,22.1]$ \\
M3B & AIME 2025 & $13.1\ [9.9,16.2]$ & $25.0\ [17.8,32.2]$ & $18.3\ [12.9,23.8]$ & $16.1\ [9.5,22.8]$ \\
Q7B & MATH-87 & $78.2\ [70.6,85.7]$ & $78.9\ [73.0,84.9]$ & $75.9\ [70.9,80.8]$ & $76.2\ [67.1,85.4]$ \\
Q7B & AIME 2024 & $11.9\ [8.8,15.1]$ & $17.5\ [13.4,21.6]$ & $15.6\ [11.2,19.9]$ & $12.8\ [8.5,17.1]$ \\
Q7B & AIME 2025 & $10.0\ [4.5,15.5]$ & $17.5\ [15.4,19.6]$ & $16.7\ [12.5,20.8]$ & $6.9\ [2.6,11.3]$ \\
M8B & MATH-87 & $83.1\ [79.8,86.4]$ & $83.9\ [78.2,89.6]$ & $84.3\ [77.7,90.9]$ & $79.7\ [73.1,86.3]$ \\
M8B & AIME 2024 & $35.6\ [32.4,38.7]$ & $43.1\ [33.7,52.4]$ & $39.4\ [35.1,43.8]$ & $31.7\ [19.2,44.1]$ \\
M8B & AIME 2025 & $27.2\ [22.0,32.4]$ & $29.7\ [28.5,30.9]$ & $31.1\ [24.5,37.8]$ & $24.2\ [18.7,29.6]$ \\
L8B & MATH-87 & $41.0\ [31.8,50.2]$ & $47.9\ [38.7,57.1]$ & $27.6^{\dagger}$ & $32.2^{\dagger}$ \\
L8B & AIME 2024 & $5.0\ [1.4,8.6]$ & $3.1\ [0.7,5.4]$ & $2.5^{\dagger}$ & $4.2^{\dagger}$ \\
L8B & AIME 2025 & $0.8\ [-1.2,2.9]$ & $1.7\ [-0.4,3.7]$ & $0.0^{\dagger}$ & $0.0^{\dagger}$ \\
Q14B & MATH-87 & $74.3\ [64.3,84.4]$ & $78.2\ [72.5,83.9]$ & $75.5\ [66.8,84.2]$ & $75.5\ [69.5,81.4]$ \\
Q14B & AIME 2024 & $14.2\ [14.2,14.2]$ & $19.4\ [16.3,22.6]$ & $16.9\ [14.6,19.3]$ & $18.9\ [13.7,24.1]$ \\
Q14B & AIME 2025 & $13.3\ [6.2,20.5]$ & $17.2\ [14.8,19.6]$ & $13.1\ [11.9,14.3]$ & $16.7\ [14.6,18.7]$ \\
M14B & MATH-87 & $86.2\ [83.4,89.1]$ & $89.3\ [87.6,90.9]$ & $88.5\ [85.7,91.4]$ & $87.7\ [86.1,89.4]$ \\
M14B & AIME 2024 & $48.1\ [45.7,50.4]$ & $47.5\ [45.4,49.6]$ & $50.0\ [46.4,53.6]$ & $37.5\ [29.2,45.8]$ \\
M14B & AIME 2025 & $31.4\ [27.1,35.7]$ & $35.6\ [23.1,48.0]$ & $35.3\ [32.9,37.7]$ & $27.8\ [22.6,33.0]$ \\
\bottomrule
\end{tabular}}
\end{center}
\vfill
\clearpage

\vspace*{\fill}
\begin{center}
\begin{minipage}{\textwidth}
\captionof{table}{\textbf{Region grafts evaluated on held-out mathematics problems.} Accuracy (\%) of the ungrafted host, its region graft and the reasoning-trained model on the 413 MATH-500 problems not used to identify the span. The 87 selection problems were answered in the same evaluation for comparison. Each condition generated one response per problem in a single generation run, using the single-user-turn prompt described in Methods. These are descriptive estimates. The MATH-87 span was fixed before evaluating the held-out problems. Model codes follow Fig.~\ref{fig:graft}.}
\label{tab:ed_heldout}
\centering
{\small\setlength{\tabcolsep}{7pt}
\begin{tabular}{lrrrrr}
\toprule
\rowcolor{gray!15} & \multicolumn{2}{c}{Selection problems (87)} & \multicolumn{3}{c}{Held-out problems (413)} \\
\rowcolor{gray!15}Host & No graft & Region & No graft & Region & Reasoning-trained \\
\midrule
M3B  & 80.5 & 81.6 & 85.5 & 88.6 & 87.9 \\
Q7B  & 77.0 & 78.2 & 83.5 & 85.5 & 93.0 \\
M8B  & 85.1 & 86.2 & 89.6 & 90.3 & 86.2 \\
L8B  & 36.8 & 46.0 & 50.1 & 53.8 & 86.0 \\
Q14B & 66.7 & 81.6 & 80.4 & 82.6 & 93.7 \\
M14B & 89.7 & 89.7 & 92.0 & 91.0 & 90.6 \\
\bottomrule
\end{tabular}
}
\end{minipage}
\end{center}
\vfill\clearpage

\vspace*{\fill}
\begin{center}
\begin{minipage}{\textwidth}
\captionof{table}{\textbf{Graft gains compared with forced continuation.} Accuracy (\%) of five instruction-tuned hosts without a graft, with forced continuation towards the shorter or longer target length, and with the region graft. Panels show \textbf{a}, MATH-87; \textbf{b}, AIME~2024; and \textbf{c}, AIME~2025. Set lengths are target generated-token counts; responses could stop below the target (Methods). Values are means and 95\% confidence intervals across three generation runs (MATH-87, 87 responses per run; AIME, 120 responses per run). The four conditions were generated together and are compared within this evaluation. Q7B was not forced (Methods). Model codes follow Fig.~\ref{fig:graft}.}
\label{tab:ed_forcing}
\centering
{\footnotesize\setlength{\tabcolsep}{4pt}
\begin{tabular}{llrrrr}
\multicolumn{6}{l}{\textbf{a}\quad\textbf{MATH-87}}\\[3pt]
\toprule
\rowcolor{gray!15}Host & Set lengths & No graft & Forced, shorter & Forced, longer & Region \\
\midrule
M3B  & 4{,}300 / 8{,}000 & $78.5\ [76.9,80.2]$ & $79.7\ [75.3,84.1]$ & $80.8\ [76.5,85.2]$ & $82.4\ [80.7,84.0]$ \\
M8B  & 1{,}200 / 4{,}000 & $82.0\ [78.7,85.3]$ & $85.8\ [79.9,91.8]$ & $82.0\ [72.8,91.2]$ & $84.7\ [80.3,89.0]$ \\
L8B  & 2{,}200 / 4{,}000 & $39.9\ [30.7,49.0]$ & $37.5\ [33.2,41.9]$ & $36.4\ [32.0,40.8]$ & $45.6\ [39.0,52.2]$ \\
Q14B & 800 / 4{,}000 & $70.9\ [67.6,74.2]$ & $69.3\ [60.2,78.5]$ & $64.8\ [57.6,71.9]$ & $77.0\ [77.0,77.0]$ \\
M14B & 2{,}400 / 4{,}000 & $86.6\ [80.0,93.2]$ & $88.5\ [82.8,94.2]$ & $87.4\ [87.4,87.4]$ & $88.1\ [83.8,92.5]$ \\
\bottomrule
\end{tabular}\par\vspace{12pt}
\begin{tabular}{llrrrr}
\multicolumn{6}{l}{\textbf{b}\quad\textbf{AIME 2024}}\\[3pt]
\toprule
\rowcolor{gray!15}Host & Set lengths & No graft & Forced, shorter & Forced, longer & Region \\
\midrule
M3B  & 10{,}100 / 12{,}000 & $25.8\ [18.7,33.0]$ & $29.2\ [18.2,40.1]$ & $27.5\ [23.9,31.1]$ & $38.3\ [32.1,44.5]$ \\
M8B  & 7{,}400 / 11{,}100 & $36.1\ [27.7,44.5]$ & $38.1\ [32.1,44.0]$ & $38.9\ [37.7,40.1]$ & $43.1\ [32.6,53.5]$ \\
L8B  & 3{,}000 / 4{,}500 & $5.8\ [0.3,11.3]$ & $4.2\ [-1.3,9.7]$ & $4.2\ [0.0,8.3]$ & $3.6\ [-2.4,9.6]$ \\
Q14B & 1{,}800 / 2{,}700 & $13.6\ [9.3,17.9]$ & $11.1\ [8.7,13.5]$ & $10.0\ [4.5,15.5]$ & $22.2\ [15.9,28.6]$ \\
M14B & 7{,}800 / 11{,}700 & $45.0\ [32.1,57.9]$ & $41.7\ [32.2,51.1]$ & $43.6\ [41.2,46.0]$ & $51.9\ [48.8,55.1]$ \\
\bottomrule
\end{tabular}\par\vspace{12pt}
\begin{tabular}{llrrrr}
\multicolumn{6}{l}{\textbf{c}\quad\textbf{AIME 2025}}\\[3pt]
\toprule
\rowcolor{gray!15}Host & Set lengths & No graft & Forced, shorter & Forced, longer & Region \\
\midrule
M3B  & 9{,}700 / 12{,}000 & $20.3\ [17.1,23.4]$ & $23.1\ [19.9,26.2]$ & $22.8\ [17.6,28.0]$ & $27.2\ [17.7,36.8]$ \\
M8B  & 8{,}100 / 12{,}000 & $29.7\ [23.4,36.0]$ & $30.6\ [25.8,35.3]$ & $30.6\ [25.3,35.8]$ & $33.9\ [25.3,42.5]$ \\
L8B  & 3{,}700 / 5{,}550 & $0.0\ [0.0,0.0]$ & $0.0\ [0.0,0.0]$ & $0.0\ [0.0,0.0]$ & $1.4\ [0.2,2.6]$ \\
Q14B & 1{,}600 / 2{,}400 & $12.8\ [5.5,20.0]$ & $8.1\ [0.8,15.3]$ & $8.1\ [6.9,9.3]$ & $16.7\ [14.6,18.7]$ \\
M14B & 9{,}000 / 12{,}000 & $32.2\ [29.1,35.4]$ & $33.3\ [27.9,38.8]$ & $33.9\ [32.7,35.1]$ & $40.0\ [32.8,47.2]$ \\
\bottomrule
\end{tabular}\par\vspace{12pt}
}
\end{minipage}
\end{center}
\vfill\clearpage

\vspace*{\fill}
\begin{lrbox}{\EDbox}%
{\footnotesize
\newcommand{\rrn}[3]{#1 & #2 & \ensuremath{#3}}
\setlength{\tabcolsep}{3pt}
\begin{tabular}{l*{4}{rrr}}
\toprule
\rowcolor{gray!15}Host & \multicolumn{3}{c}{Region} & \multicolumn{3}{c}{None} & \multicolumn{3}{c}{Low} & \multicolumn{3}{c}{High} \\
\cmidrule(lr){2-4}\cmidrule(lr){5-7}\cmidrule(lr){8-10}\cmidrule(lr){11-13}
 & \shortstack{Retained\\(\%)} & \shortstack{Recovered\\(\%)} & \shortstack{Net\\(no.)} & \shortstack{Retained\\(\%)} & \shortstack{Recovered\\(\%)} & \shortstack{Net\\(no.)} & \shortstack{Retained\\(\%)} & \shortstack{Recovered\\(\%)} & \shortstack{Net\\(no.)} & \shortstack{Retained\\(\%)} & \shortstack{Recovered\\(\%)} & \shortstack{Net\\(no.)} \\
\midrule
\multicolumn{13}{l}{\textbf{a}\quad MATH-87}\\
M3B  & \rrn{94}{40}{+3.3} & \rrn{77}{19}{-12.5} & \rrn{83}{16}{-8.5} & \rrn{94}{40}{+3.5} \\
Q7B  & \rrn{90}{39}{+0.7} & \rrn{89}{31}{-1.9}  & \rrn{86}{29}{-4.2} & \rrn{90}{35}{-0.3} \\
M8B  & \rrn{95}{30}{+0.7} & \rrn{89}{28}{-3.9}  & \rrn{94}{44}{+1.8} & \rrn{94}{31}{-0.2} \\
L8B  & \rrn{82}{24}{+6.0} & \rrn{58}{11}{-9.4}  & \rrn{78}{21}{+2.8} & \rrn{73}{20}{+0.9} \\
Q14B & \rrn{93}{36}{+3.3} & \rrn{87}{27}{-2.1}  & \rrn{88}{30}{-0.9} & \rrn{93}{34}{+3.2} \\
M14B & \rrn{96}{44}{+2.7} & \rrn{95}{33}{+0.1}  & \rrn{97}{43}{+2.8} & \rrn{96}{42}{+2.2} \\
\midrule
\multicolumn{13}{l}{\textbf{b}\quad AIME 2024}\\
M3B  & \rrn{88}{22}{+3.7} & \rrn{48}{2}{-3.9}  & \rrn{65}{14}{+0.0} & \rrn{86}{21}{+3.3} \\
Q7B  & \rrn{82}{8}{+1.3}  & \rrn{85}{3}{+0.1}  & \rrn{96}{3}{+0.8}  & \rrn{86}{3}{+0.2} \\
M8B  & \rrn{92}{24}{+3.3} & \rrn{73}{13}{-0.9} & \rrn{90}{23}{+3.0} & \rrn{86}{17}{+1.3} \\
L8B  & \rrn{17}{0}{-1.0}  & \rrn{3}{0}{-1.2}   & \rrn{58}{3}{+0.2}  & \rrn{42}{1}{-0.7} \\
Q14B & \rrn{93}{11}{+2.3} & \rrn{72}{1}{-1.1}  & \rrn{63}{4}{-0.9}  & \rrn{83}{5}{+0.5} \\
M14B & \rrn{89}{15}{+0.0} & \rrn{74}{13}{-2.9} & \rrn{82}{21}{-0.5} & \rrn{85}{21}{+0.0} \\
\midrule
\multicolumn{13}{l}{\textbf{c}\quad AIME 2025}\\
M3B  & \rrn{100}{17}{+4.3} & \rrn{64}{7}{+0.2} & \rrn{77}{9}{+1.3}  & \rrn{92}{14}{+3.2} \\
Q7B  & \rrn{100}{11}{+3.0} & \rrn{51}{4}{-0.4} & \rrn{69}{5}{+0.5}  & \rrn{69}{6}{+0.8} \\
M8B  & \rrn{88}{12}{+1.7}  & \rrn{86}{6}{+0.1} & \rrn{94}{11}{+2.0} & \rrn{90}{12}{+1.8} \\
L8B  & \rrn{--}{0}{+0.0}   & \rrn{--}{1}{+0.2} & \rrn{--}{2}{+0.5}  & \rrn{--}{0}{+0.1} \\
Q14B & \rrn{82}{7}{+0.7}   & \rrn{73}{1}{-1.1} & \rrn{78}{3}{-0.4}  & \rrn{84}{6}{+0.7} \\
M14B & \rrn{83}{20}{+2.3}  & \rrn{83}{10}{+0.3} & \rrn{91}{11}{+1.5} & \rrn{86}{15}{+1.8} \\
\bottomrule
\end{tabular}}%
\end{lrbox}
\begin{center}
\begin{minipage}{\textwidth}
\captionof{table}{\textbf{Retention, recovery and net change by benchmark and region overlap.}
Columns report mean retention, recovery and net change for the MATH-87-defined region graft or equally wide grafts with none, low or high region overlap on \textbf{a}, MATH-87; \textbf{b}, AIME 2024; and \textbf{c}, AIME 2025.
Overlap is the fraction of region layers included in a graft: none, $0$; low, $(0,0.5)$; high, $[0.5,1)$; the region graft has full overlap.
Retention is the percentage of problems solved by the ungrafted instruction-tuned model that remain solved after grafting; recovery is the percentage of previously unsolved problems that become solved.
A problem is solved when its single MATH-87 response or at least two of its four AIME responses are correct.
Transition measures were first averaged across evaluated graft positions within each model, benchmark, overlap class and generation run, then across the three runs. Retention and recovery are rounded to the nearest percentage point; net is the mean number fixed minus broken. An em dash indicates that the no-graft host solved no problem, leaving retention undefined. Model codes follow Fig.~\ref{fig:graft}.}
\label{tab:ed_census_detail}
\centering
\usebox{\EDbox}
\end{minipage}
\end{center}
\vfill
\clearpage

\vspace*{\fill}
\begin{center}
\captionof{table}{\textbf{Problem-level gains of region grafts on AIME and reproduction of the donor's wrong answers.} \textbf{a}, Problems the host did not solve (its modal answer over 12 responses, three generation runs $\times$ four, was incorrect) are split by whether the donor solved them (its modal answer over nine responses was correct); $n$ is the number of problems in each class, and the two $n$ sum to the host's unsolved problems out of 30. Accuracy change is the mean over the problems in a class of the change in the fraction of correct responses between the region graft and the host; --, no problems in the class. \textbf{b}, Own donor minus other donors is, for the region graft, the Spearman correlation between problem-level gain and the own donor's accuracy profile minus the mean correlation with the other five donors, over the problems the host did not always solve (Methods). Wrong answers reproduced is the mean over the $n$ eligible problems of the percentage of 12 host or region-graft responses matching the own donor's modal wrong answer; eligible problems are those where the donor's modal answer is parseable, incorrect and differs from the host's modal answer. Model codes follow Fig.~\ref{fig:graft}; these comparisons are descriptive.}
\label{tab:ed_policy_problems}
\par\vspace{7pt}
{\small\setlength{\tabcolsep}{6pt}
\begin{tabularx}{\textwidth}{ll*{2}{>{\raggedleft\arraybackslash}X}}
\multicolumn{4}{l}{\textbf{a}\quad\textbf{Gains on problems the host did not solve}}\\[3pt]
\toprule
\rowcolor{gray!15}Host & Benchmark & \shortstack{Accuracy change (pp),\\donor solved ($n$)} & \shortstack{Accuracy change (pp),\\donor failed ($n$)} \\
\midrule
M3B & AIME 2024 & -- (0) & $+4.7$ (16) \\
M3B & AIME 2025 & $+41.7$ (4) & $+4.6$ (20) \\
Q7B & AIME 2024 & $+6.8$ (16) & $+0.9$ (9) \\
Q7B & AIME 2025 & $+10.6$ (11) & $+1.1$ (15) \\
M8B & AIME 2024 & -- (0) & $+5.7$ (16) \\
M8B & AIME 2025 & $+25.0$ (1) & $+4.6$ (18) \\
L8B & AIME 2024 & $+1.0$ (17) & 0.0 (10) \\
L8B & AIME 2025 & $+1.2$ (14) & 0.0 (16) \\
Q14B & AIME 2024 & $+3.7$ (18) & 0.0 (6) \\
Q14B & AIME 2025 & $+9.8$ (11) & $+2.1$ (12) \\
M14B & AIME 2024 & $+12.5$ (2) & $+1.0$ (8) \\
M14B & AIME 2025 & $-8.3$ (1) & $+1.3$ (13) \\
\bottomrule
\end{tabularx}}
\par\vspace{14pt}
{\small\setlength{\tabcolsep}{4pt}
\begin{tabularx}{\textwidth}{ll*{4}{>{\raggedleft\arraybackslash}X}}
\multicolumn{6}{l}{\textbf{b}\quad\textbf{Donor specificity and reproduction of wrong answers}}\\[3pt]
\toprule
\rowcolor{gray!15}Host & Benchmark & \shortstack{Own donor minus\\other donors\\($\Delta\rho$)} & \shortstack{Eligible\\problems ($n$)} & \shortstack{Wrong answers\\reproduced, host\\(\%)} & \shortstack{Wrong answers\\reproduced, region\\(\%)} \\
\midrule
M3B & AIME 2024 & $-0.04$ & 17 & 2.5 & 3.4 \\
M3B & AIME 2025 & $-0.05$ & 18 & 2.3 & 5.1 \\
Q7B & AIME 2024 & $-0.09$ & 8 & 2.1 & 8.3 \\
Q7B & AIME 2025 & $-0.01$ & 13 & 1.9 & 6.4 \\
M8B & AIME 2024 & $-0.15$ & 18 & 1.4 & 0.9 \\
M8B & AIME 2025 & $+0.08$ & 17 & 0.5 & 1.0 \\
L8B & AIME 2024 & $-0.18$ & 10 & 1.7 & 0.8 \\
L8B & AIME 2025 & $0.00$ & 16 & 0.0 & 1.0 \\
Q14B & AIME 2024 & $-0.08$ & 6 & 0.0 & 1.4 \\
Q14B & AIME 2025 & $-0.29$ & 10 & 2.5 & 4.2 \\
M14B & AIME 2024 & $-0.19$ & 7 & 1.2 & 0.0 \\
M14B & AIME 2025 & $+0.02$ & 15 & 1.1 & 2.2 \\
\bottomrule
\end{tabularx}}
\end{center}
\vfill\clearpage

\setcounter{figure}{0}
\renewcommand{\figurename}{Extended Data Fig.}
\renewcommand{\thefigure}{\arabic{figure}}
\providecommand{\theHfigure}{}
\renewcommand{\theHfigure}{ED.\arabic{figure}}

\vspace*{\fill}
\begin{center}
\begin{overpic}[width=\textwidth]{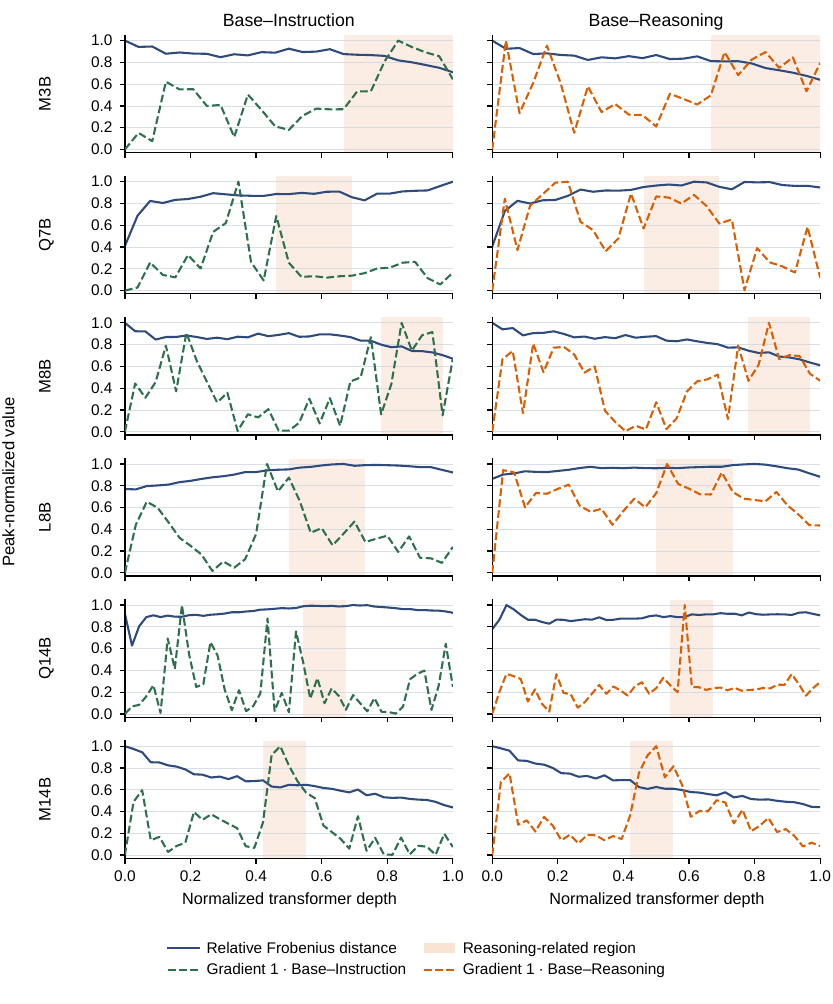}
\put(3,98){\textbf{\sffamily a}}\put(47.5,98){\textbf{\sffamily b}}
\end{overpic}
\captionof{figure}{\textbf{Layer-wise gradient displacement is not consistently aligned with parameter-change magnitude.}
Relative Frobenius distance (navy solid line) and unsmoothed absolute gradient~1 displacement on MATH-87 inputs (dashed lines) are shown for base--instruction (\textbf{a}, green) and base--reasoning (\textbf{b}, orange) comparisons across six matched model sets. The final transformer layer is excluded. Each profile is divided by its own maximum to compare profile shapes. Shading marks the same layer span selected from MATH-87 base--reasoning displacement in both columns. Model codes follow Fig.~\ref{fig:detection}.}
\label{fig:ed_weight_grad1}
\end{center}
\vfill
\clearpage

\vspace*{\fill}
\begin{center}
\includegraphics[width=\textwidth]{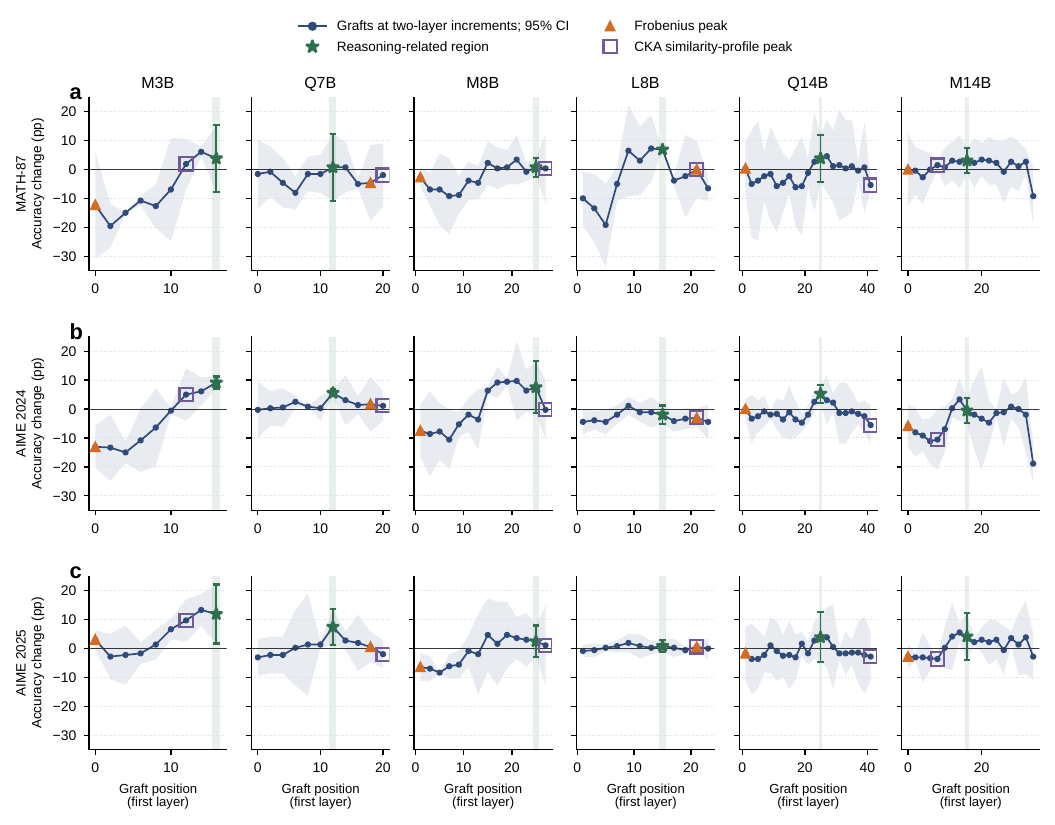}
\par\vspace{8pt}
{\small\setlength{\tabcolsep}{6pt}\renewcommand{\arraystretch}{1.15}
\begin{tabular}{lrrrrrr}
\multicolumn{7}{l}{\textbf{d}\quad Selected layer spans}\\[3pt]
\toprule
\rowcolor{gray!15}Selection & M3B & Q7B & M8B & L8B & Q14B & M14B \\
\midrule
Region & 16--24 & 12--18 & 25--31 & 15--22 & 25--31 & 16--21 \\
Frobenius & 0--8 & 18--24 & 1--7 & 21--28 & 1--7 & 0--5 \\
CKA-row & 12--20 & 20--26 & 27--33 & 21--28 & 41--47 & 8--13 \\
\bottomrule
\end{tabular}}
\captionof{figure}{\textbf{Graft effects across layer positions and alternative selection profiles.}
\textbf{a--c}, Accuracy change from the ungrafted instruction-tuned model for MATH-87, AIME~2024 and AIME~2025, with models arranged by column.
Each graft copied reasoning-trained weights into the indicated span without further training, with width held fixed within each model.
The horizontal coordinate gives the first layer of each graft (zero-based).
Navy points and lines show all 85 positions per benchmark used in Fig.~\ref{fig:graft}, including the six region grafts, on a grid with two-layer spacing.
Green stars mark region grafts; orange triangles and open purple squares mark equally wide grafts selected using peaks in relative Frobenius distance and CKA similarity-profile change, respectively (Methods).
The two alternatives select the same span in L8B, shown by nested symbols.
Symbols show means; shading and error bars show 95\% confidence intervals across three generation runs.
\textbf{d}, Selected layer spans (zero-based, inclusive), held fixed across benchmarks.
Model codes follow Fig.~\ref{fig:graft}.}
\label{fig:ed_stride_profiles}
\end{center}
\vfill
\clearpage

\vspace*{\fill}
\begin{center}
\includegraphics[width=\textwidth]{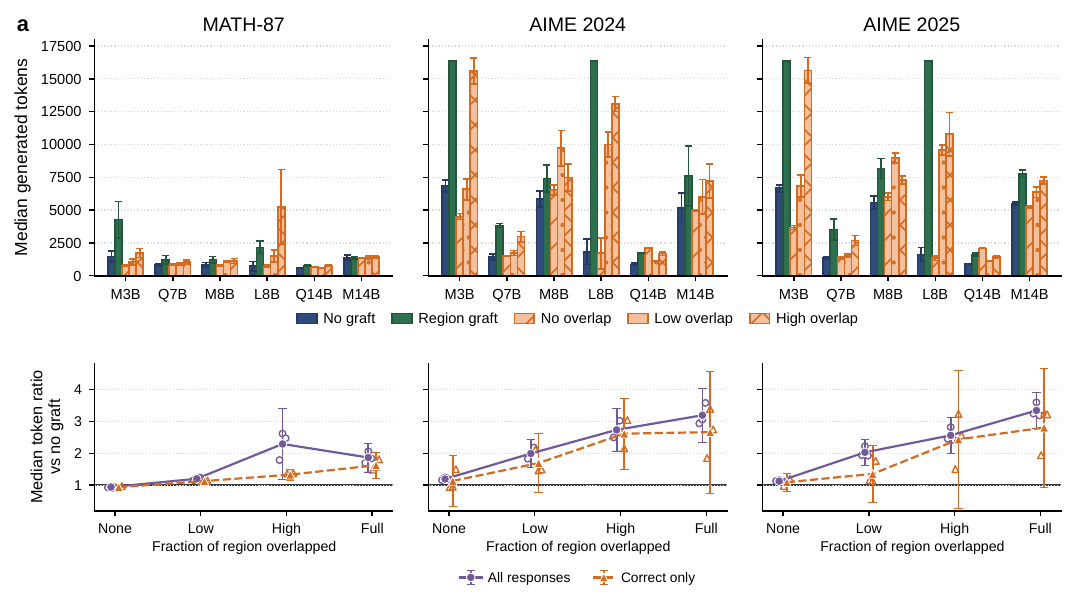}\\[6pt]
\includegraphics[width=\textwidth]{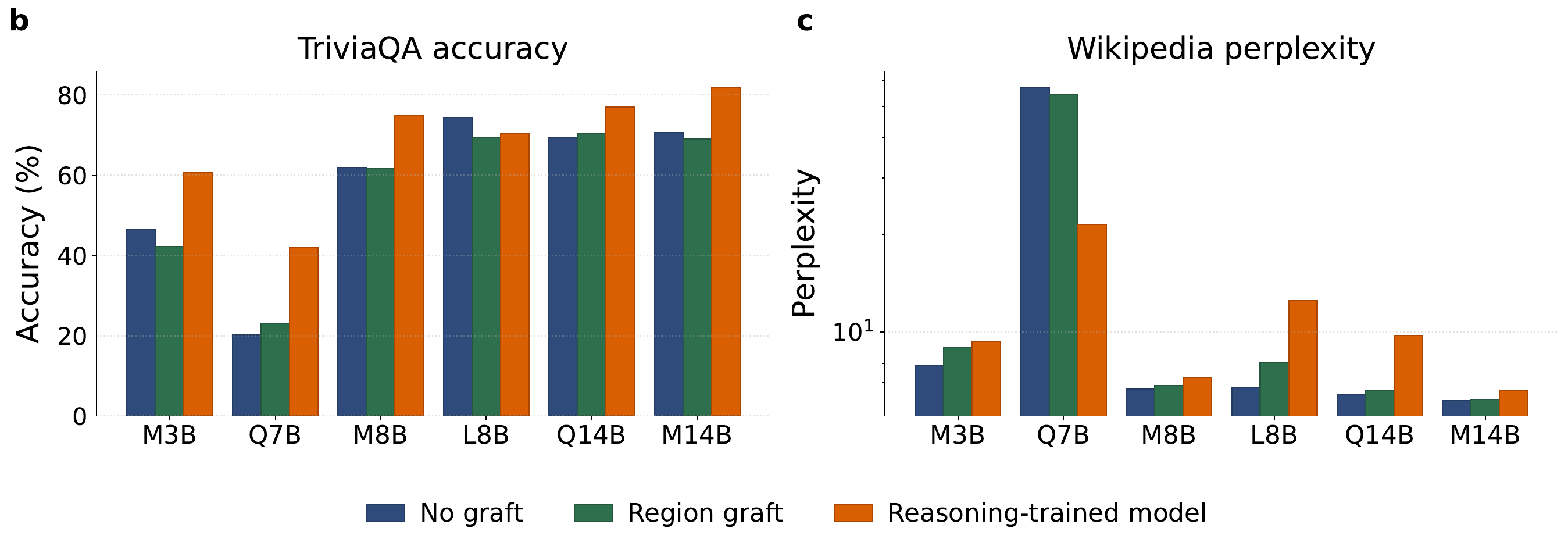}
\captionof{figure}{\textbf{Solution length, factual recall and language modelling after region grafting.}
\textbf{a}, Upper plots show median response length for the ungrafted instruction-tuned model, the MATH-87 region graft and equally wide grafts with no, low or high region overlap.
Overlap is the fraction of region layers included in a graft: none, $0$; low, $(0,0.5)$; high, $[0.5,1)$; full, $1$ (region graft).
Lower plots show ratios to the ungrafted median using all responses or only correct responses in each condition.
Within each model and generation run, medians and ratios are averaged across graft positions in each overlap class; lower-plot ratios are then averaged equally across models with defined values.
Bars and filled symbols show means across three generation runs; whiskers show 95\% confidence intervals.
Open symbols in the lower plots show individual generation runs.
\textbf{b}, TriviaQA accuracy on the same 500 questions for each model condition, using one greedily decoded response per question and alias-based scoring.
\textbf{c}, Wikipedia perplexity on the same 100 passages, calculated by exponentiating the mean passage-level negative log-likelihood; the vertical axis is logarithmic and lower values indicate better language modelling.
Panels \textbf{b,c} compare the ungrafted instruction-tuned model, region graft and reasoning-trained model in a single descriptive evaluation per condition.
Model codes follow Fig.~\ref{fig:graft}.}
\label{fig:ed_knowledge}
\end{center}
\vfill

\clearpage
\vspace*{\fill}
\begin{center}
\includegraphics[width=\textwidth]{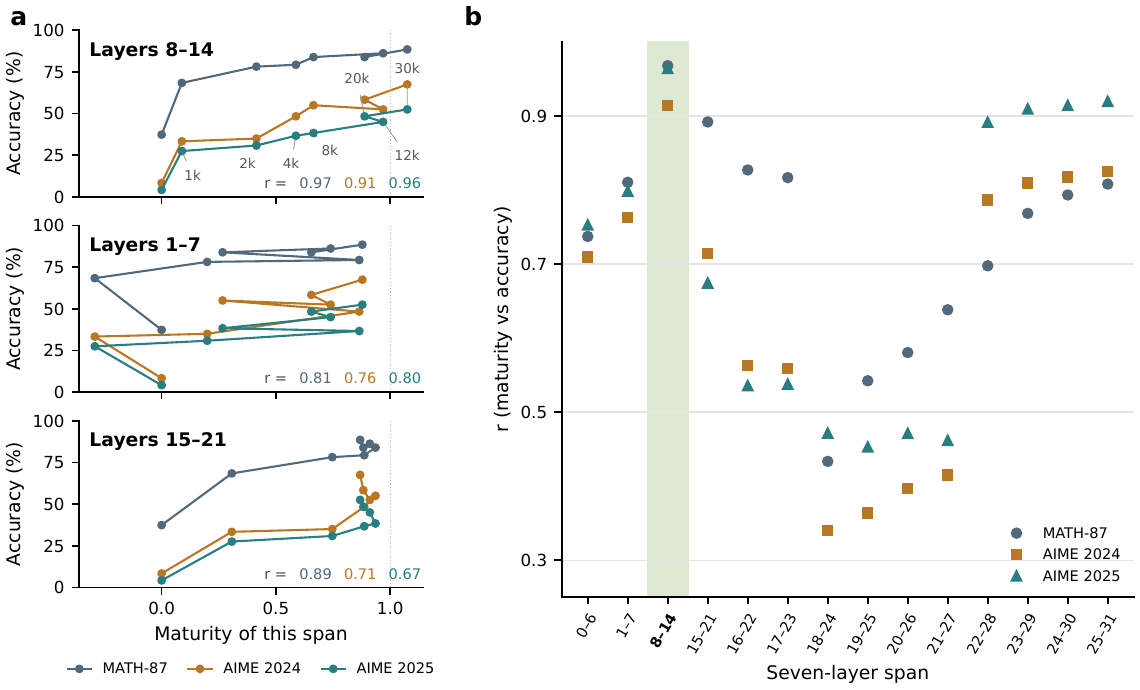}
\captionof{figure}{\textbf{Maturity of the layers 8--14 pattern and accuracy across OLMo reasoning-training stages.}
\textbf{a}, Accuracy versus maturity for the base model and seven intermediate stages, joined in training order; the endpoint is omitted and the dotted line marks maturity one. Top, layers 8--14; middle and bottom, the two equally wide non-overlapping spans flanking the region (layers 1--7 and 15--21), each scored with the same index referenced to its own endpoint pattern. The Pearson correlation in each panel uses the seven intermediate stages and excludes the base model. MATH-87 uses 87 problems, averaging two generation runs through 4k steps and one run from 8k to 30k; AIME uses four responses to each of 30 problems per year in one run (Methods).
\textbf{b}, The same correlation for layers 8--14 (shaded) and the 13 equally wide spans with no overlap, ordered by their first layer.}
\label{fig:ed_olmo_maturity}
\end{center}
\vfill\clearpage
\vspace*{\fill}
\begin{center}
\includegraphics[width=\textwidth]{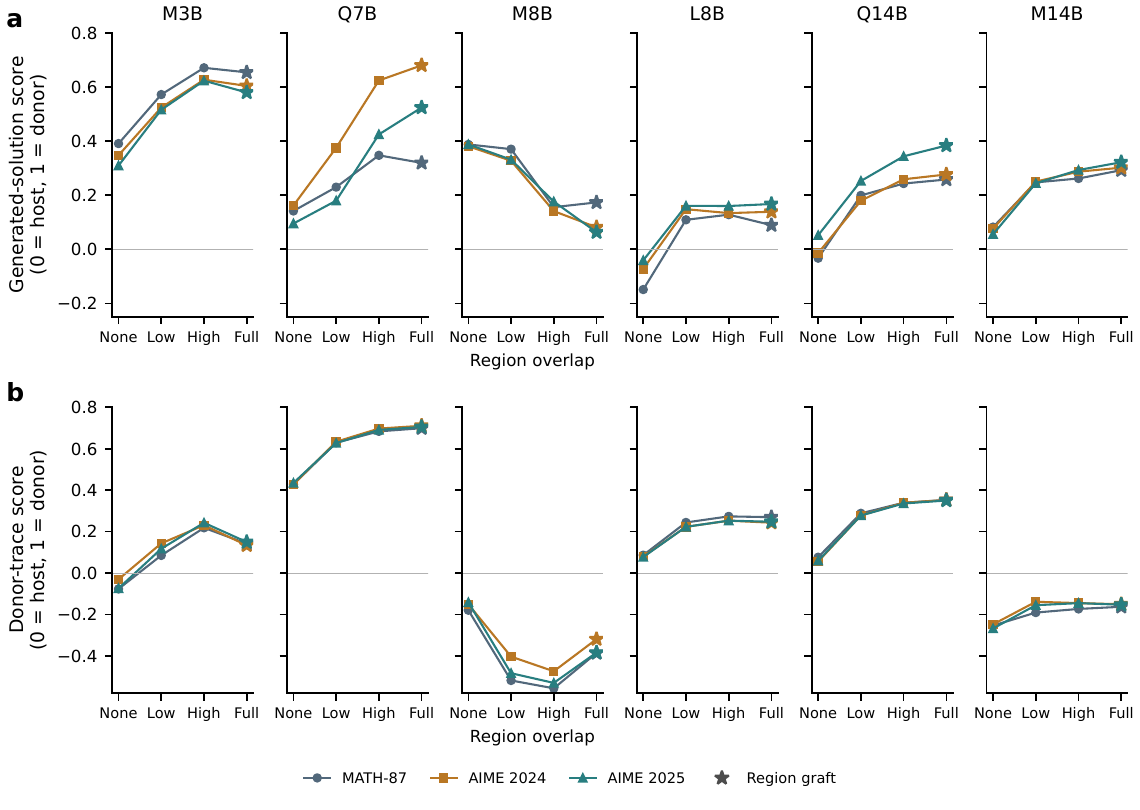}
\captionof{figure}{\textbf{Donor-likeness of grafted hosts by region overlap.}
\textbf{a}, Generated-solution score of each host's grafts (0, host; 1, donor; negative, further from the donor than the host; Methods), averaged over the graft positions in each overlap class for MATH-87, AIME 2024 and AIME 2025. Overlap classes follow Fig.~\ref{fig:graft}: none, 0; low, $(0,0.5)$; high, $[0.5,1)$; full, the single region graft (star). The grey line marks the host's own score.
\textbf{b}, Donor-trace score of the same grafts, the fraction of the host--donor divergence closed on the donor's own written solutions (0, host; 1, donor; Methods), averaged in the same way. On MATH-87, the donor solutions retained for scoring after excluding those longer than the scoring limit numbered 85, 80, 87, 76, 82 and 85 for M3B, Q7B, M8B, L8B, Q14B and M14B; on AIME all 30 were retained.
Scores use one response per problem from a single generation run; intervals are not shown (Statistics and reproducibility). Model codes follow Fig.~\ref{fig:graft}.}
\label{fig:ed_policy_positions}
\end{center}
\vfill

\end{document}


{\Large\bfseries Supplementary Information\par}
\vspace{4pt}
{\large Transferring an analysis of human brain organization to language models\par}
Donghyeok Lee, Sangwon Yu, Sangkwon Park, Hye Youn Park, Munyoung Chang and Sungroh Yoon

\notetitle{Terms and model sets}
In each of the six matched model sets, the base model is the pretrained reference, the instruction-tuned model is the host that receives a graft, and the reasoning-trained model is the donor that supplies the weights.
A graft replaces the weights of a contiguous span of host layers with the donor's weights for the same layers, without further training.
The reasoning-related region, or region, is the span selected from base--reasoning displacement on MATH-87 (Methods).
A comparison graft has the region's width at another position, and its overlap is the fraction of region layers that it contains.
A generation run is one sampling of responses for every condition; the three runs repeat the same problems.
MATH-87 is the fixed 87-problem subset of MATH-500 used to identify regions, and AIME~2024 and AIME~2025 contain 30 problems each.
CKA denotes centred kernel alignment.
Layer indices are zero-based and spans are inclusive.

Supplementary Table~\ref{tab:si_models} lists the models of each set, with the codes used throughout, and the dimensions of the sampled projections.

\captionof{table}{\textbf{Matched model sets, checkpoints and architecture.} The six sets used in the main reasoning analysis are shown. \textbf{a}, Model checkpoints and representation-node counts. \textbf{b}, Architecture and projection dimensions. UltraMedical, used in the medical analysis, shares the Llama 8B architecture (Methods). $L$, number of transformer layers; $U=11L$, number of representation nodes; $h$, residual-stream hidden size; $i$, feed-forward intermediate size; $a$ and $a_{\mathrm{KV}}$, query- and key/value-head counts; $d_{\mathrm{head}}$, head dimension; $d_q=a d_{\mathrm{head}}$; $d_{\mathrm{KV}}=a_{\mathrm{KV}}d_{\mathrm{head}}$.}
\label{tab:si_models}
{\footnotesize\setlength{\tabcolsep}{4pt}
\begin{tabularx}{\linewidth}{ll>{\raggedright\arraybackslash}Xrr}
\multicolumn{5}{l}{\textbf{a}\quad\textbf{Model checkpoints and representation-node counts}}\\[3pt]
\toprule
\rowcolor{gray!15}Model set & Variant & Model checkpoint & $L$ & $U$ \\
\midrule
Ministral 3B (M3B) & Base & Ministral-3-3B-Base-2512 & 26 & 286 \\
 & Instruction-tuned & Ministral-3-3B-Instruct-2512-BF16 &  &  \\
 & Reasoning-trained & Ministral-3-3B-Reasoning-2512 &  &  \\
\addlinespace[3pt]
Qwen 7B (Q7B) & Base & Qwen2.5-Math-7B & 28 & 308 \\
 & Instruction-tuned & Qwen2.5-Math-7B-Instruct &  &  \\
 & Reasoning-trained & DeepSeek-R1-Distill-Qwen-7B &  &  \\
\addlinespace[3pt]
Ministral 8B (M8B) & Base & Ministral-3-8B-Base-2512 & 34 & 374 \\
 & Instruction-tuned & Ministral-3-8B-Instruct-2512-BF16 &  &  \\
 & Reasoning-trained & Ministral-3-8B-Reasoning-2512 &  &  \\
\addlinespace[3pt]
Llama 8B (L8B) & Base & Llama-3.1-8B & 32 & 352 \\
 & Instruction-tuned & Llama-3.1-8B-Instruct &  &  \\
 & Reasoning-trained & DeepSeek-R1-Distill-Llama-8B &  &  \\
\addlinespace[3pt]
Qwen 14B (Q14B) & Base & Qwen2.5-14B & 48 & 528 \\
 & Instruction-tuned & Qwen2.5-14B-Instruct &  &  \\
 & Reasoning-trained & DeepSeek-R1-Distill-Qwen-14B &  &  \\
\addlinespace[3pt]
Ministral 14B (M14B) & Base & Ministral-3-14B-Base-2512 & 40 & 440 \\
 & Instruction-tuned & Ministral-3-14B-Instruct-2512-BF16 &  &  \\
 & Reasoning-trained & Ministral-3-14B-Reasoning-2512 &  &  \\
\bottomrule
\end{tabularx}}
\vspace{10pt}

{\footnotesize\setlength{\tabcolsep}{4pt}
\begin{tabularx}{\linewidth}{l*{7}{>{\raggedleft\arraybackslash}X}}
\multicolumn{8}{l}{\textbf{b}\quad\textbf{Architecture and projection dimensions}}\\[3pt]
\toprule
\rowcolor{gray!15}Model set & $h$ & $i$ & $a$ & $a_{\mathrm{KV}}$ & $d_{\mathrm{head}}$ & $d_q$ & $d_{\mathrm{KV}}$ \\
\midrule
Ministral 3B (M3B)  & 3{,}072 & 9{,}216  & 32 & 8 & 128 & 4{,}096 & 1{,}024 \\
Qwen 7B (Q7B)       & 3{,}584 & 18{,}944 & 28 & 4 & 128 & 3{,}584 & 512 \\
Ministral 8B (M8B)  & 4{,}096 & 14{,}336 & 32 & 8 & 128 & 4{,}096 & 1{,}024 \\
Llama 8B (L8B)      & 4{,}096 & 14{,}336 & 32 & 8 & 128 & 4{,}096 & 1{,}024 \\
Qwen 14B (Q14B)      & 5{,}120 & 13{,}824 & 40 & 8 & 128 & 5{,}120 & 1{,}024 \\
Ministral 14B (M14B) & 5{,}120 & 16{,}384 & 32 & 8 & 128 & 4{,}096 & 1{,}024 \\
\bottomrule
\end{tabularx}}

\clearpage
\notetitle{Reference and threshold dependence}
\textbf{Reference model.}
The main analysis measures displacement between the base and reasoning-trained models, capturing organizational differences accumulated during post-training.
Using the instruction-tuned model as reference instead measures what differs between the two post-trained models.
Neither reference isolates changes specific to reasoning training.
We repeated the alignment and the selection rule with this reference, changing nothing else (Supplementary Table~\ref{tab:si_reference}).

The two references selected overlapping spans in Q7B, L8B and Q14B and disjoint spans in the three Ministral sets, where the instruction-referenced spans lay within layers 0--13.
The positional evaluation includes region-width grafts overlapping these instruction-referenced spans: layers 0--8 in M3B (containing layers 1--4), layers 1--7 in M8B (four of the five layers 0--4) and layers 8--13 in M14B (six of the seven layers 7--13; Supplementary Table~\ref{tab:si_selectors}).
These evaluated windows differ in width or position from the instruction-referenced spans, which were not evaluated directly.
In all nine host--benchmark comparisons, these grafts had lower mean accuracy than the region graft, by 1.5 to 22.2 percentage points.

\captionof{table}{\textbf{Spans selected using a pretrained or instruction-tuned reference.} Both comparisons use MATH-87 inputs, omit the final transformer layer, smooth displacement with an edge-padded three-layer moving average, and use the threshold and minimum-width settings specified in Methods, held fixed within each model across the two references. Overlap gives the number of layers shared by the two spans divided by the width of the base-referenced span. Layer indices are zero-based and inclusive.}
\label{tab:si_reference}
\begin{center}\small
\begin{tabular}{lllr}
\toprule
\rowcolor{gray!15}Model & Base reference & Instruction reference & Overlap \\
\midrule
\datainput{reference_rows.tex}
\bottomrule
\end{tabular}
\end{center}

\textbf{Threshold.}
Supplementary Table~\ref{tab:si_threshold} varies the threshold with all other selection settings fixed; ties between equally long qualifying runs are resolved in favour of the lower layer indices.
The selected span shifted to a separate peak in Q7B at $f\geq0.80$ and in M8B at $f=0.60$.
L8B showed the largest change in width, from 28 layers at $f=0.60$ to five at $f=0.90$.
All seven selections for each of Q14B and M14B had the minimum width; their starting layers varied by zero and one layer, respectively.

\captionof{table}{\textbf{Selection sensitivity to the displacement threshold.} Each entry gives the selected layer span for the indicated fraction $f$ of the maximum smoothed displacement. Inputs, reference, smoothing and the 15\% minimum-width rule are held fixed. The main analysis uses $f=0.70$ except for L8B, which uses $f=0.85$.}
\label{tab:si_threshold}
\begin{center}\small
\begin{tabular}{lrrrrrrr}
\toprule
\rowcolor{gray!15}Model & .60 & .65 & .70 & .75 & .80 & .85 & .90 \\
\midrule
\datainput{threshold_rows.tex}
\bottomrule
\end{tabular}
\end{center}

\clearpage
\notetitle{Alternative span selection}
Supplementary Table~\ref{tab:si_selectors} gives the spans and accuracies behind the selector comparison in the main article and adds a central-window baseline.
Each alternative has the region's width and was selected as described in Methods, without reference to graft accuracy.
For the central window, the starting layer before grid adjustment is $\lfloor(L-1-w)/2\rfloor$, where $L-1$ is the number of analysed layers and $w$ is the region width.

The region graft had higher mean accuracy than the Frobenius-selected graft and than the CKA-row-selected graft in all 18 host--benchmark comparisons, by a mean of 7.1 percentage points (range 0.3 to 22.2) and 4.9 points (0.3 to 10.8), respectively.
The central window coincided with the region in M14B; in the 15 remaining comparisons the region was higher in 13, equal in one (L8B on AIME~2025) and lower in one (L8B on AIME~2024), with a mean difference of 6.8 points.

\captionof{table}{\textbf{Fixed-width alternative selections and graft accuracy.}
\textbf{a}, Evaluated spans. Each selector uses the same width as the host's region. Frobenius and CKA-row denote the two profiles compared in the main article; middle denotes the central-window baseline defined in the accompanying Supplementary Note.
\textbf{b}, Accuracy (\%), averaged across three generation runs of the positional evaluation. Each run contains 87 MATH-87 responses or 120 responses on each AIME benchmark (30 problems, four responses per problem). The no-graft host is included as a reference. Cells with identical spans report the same evaluation.}
\label{tab:si_selectors}
\begin{center}\small
\begin{tabular}{lllll}
\multicolumn{5}{l}{\textbf{a}\quad Selected layer spans}\\[3pt]
\toprule
\rowcolor{gray!15}Host & Region & Frobenius & CKA-row & Middle \\
\midrule
\datainput{selector_span_rows.tex}
\bottomrule
\end{tabular}

\vspace{14pt}
{\small\setlength{\tabcolsep}{4pt}
\begin{tabular}{llrrrrr}
\multicolumn{7}{l}{\textbf{b}\quad Accuracy (\%)}\\[3pt]
\toprule
\rowcolor{gray!15}Host & Benchmark & No graft & Region & Frobenius & CKA-row & Middle \\
\midrule
\datainput{selector_accuracy_rows.tex}
\bottomrule
\end{tabular}}
\end{center}

\clearpage
\notetitle{Evaluation units and prompt structure}
\textbf{Units.}
Accuracy is the fraction of correct sampled responses within a generation run; on AIME, a run contains four responses to each of 30 problems.
Answer transitions are counted per problem, and an AIME problem is solved when at least two of its four responses are correct, so one correct response of four contributes to accuracy but not to the solved count.
The three runs repeat the same problems; intervals across runs therefore reflect variability in generation and not in the sampling of problems.
Supplementary Table~\ref{tab:si_evaluation} lists the problems, responses and runs behind each comparison.

\textbf{Prompt.}
The mathematical instruction was:
\begin{quote}\small
Please reason step by step, and put your final answer within \texttt{\textbackslash boxed\{\}}.
\end{quote}
In the main positional evaluation, Q7B, L8B and M14B received the instruction in the system turn and the problem in the user turn, whereas M3B, M8B and Q14B received both in one user turn; placement was fixed within each host across benchmarks and conditions.
The forcing and held-out evaluations used a single user turn for every host and generated their own ungrafted and region-graft conditions.

OLMo stage grafts used a plain-text prompt (the instruction, a blank line, the problem and a final \texttt{Solution:} cue) and deterministic scoring that counted responses reaching the token or repetition limit as incorrect, whereas the six-host analysis used the adjudicated labels described in Methods.

\captionof{table}{\textbf{Evaluation units in mathematical graft comparisons.} The response count is per condition and generation run. Main and forcing intervals summarize three run-level estimates. In the held-out evaluation, the 87 selection problems were also answered in the same run for comparison. The main positional response limit is 16{,}384 tokens, except Q7B at 4{,}096; forcing uses 16{,}384 tokens for all five included hosts. Sampling, scoring and stopping rules are described in Methods and the accompanying Supplementary Note.}
\label{tab:si_evaluation}
\begin{center}\small
\begin{tabularx}{\linewidth}{>{\raggedright\arraybackslash}Xrr>{\raggedright\arraybackslash}p{36mm}}
\toprule
\rowcolor{gray!15}Comparison & Problems & Responses/run & Generation runs \\
\midrule
Main positions, MATH-87 & 87 & 87 & 3 \\
Main positions, each AIME year & 30 & 120 & 3 \\
Benchmark-specific selection, MATH-87 & 87 & 348 & 1 \\
Benchmark-specific selection, each AIME year & 30 & 120 & 1 \\
Forcing, MATH-87 & 87 & 87 & 3 \\
Forcing, each AIME year & 30 & 120 & 3 \\
Held-out MATH-500 & 413 & 413 & 1 \\
OLMo position $\times$ donor stage, MATH-87 & 87 & 87 & 3 \\
OLMo 8--14 stage grafts, each AIME year & 30 & 120 & 1 \\
\bottomrule
\end{tabularx}
\end{center}

\clearpage
\notetitle{OLMo stage accuracy and graft positions across training}

\textbf{Accuracy after regional maturity approached one.}
In evaluations of the ungrafted model at each training stage (Methods), AIME~2024 correct responses rose from 63 to 85 of 120 between 12{,}000 steps and the endpoint, and AIME~2025 from 54 to 70 of 120 (18.3 and 13.3 percentage points), while regional maturity remained near one (main Fig.~4c).

\textbf{Graft positions.}
The main trajectory follows layers 8--14 across eight donor stages.
Supplementary Table~\ref{tab:si_olmo_map} extends it to 13 seven-layer grafts starting at layers 0, 2, \ldots, 24, each evaluated on MATH-87 in three generation runs with the same pretrained host (312 graft evaluations); the layers 8--14 row is the series shown in main Fig.~4d.
Extended Data Fig.~4b compares maturity in layers 8--14 with that in 13 equally wide spans that do not overlap layers 8--14.

Graft accuracy varied with both position and donor stage.
For layers 8--14, the mean correct count peaked at 47.3 of 87 at 20{,}000 steps and was 41.7 at the endpoint, where layers 22--28 yielded 44.7.
Grafts at layers 24--30 declined from 42.3 correct responses at 8{,}000 steps to 12.3 at the endpoint.

Supplementary Table~\ref{tab:si_olmo_transitions} separates recovered and broken answers at the endpoint.
At the endpoint, retention was 75.9\% for layers 22--28 and 72.0\% for layers 8--14.

\captionof{table}{\textbf{Complete OLMo graft-position by donor-stage map.} Entries are mean correct-response counts out of 87 MATH-87 problems across three generation runs. The ungrafted host answered 30, 39 and 35 problems correctly in the first, second and third generation runs, respectively (mean 34.7/87). Each graft replaces seven layers; all other host parameters remain fixed. Stage labels are supervised fine-tuning steps, with End denoting the final stage. The 8--14 row reproduces the main fixed-span comparison. Responses reaching the token or repetition limit count as incorrect.}
\label{tab:si_olmo_map}
\begin{center}\small
\begin{tabular}{lrrrrrrrr}
\toprule
\rowcolor{gray!15}Layers & 1k & 2k & 4k & 8k & 12k & 20k & 30k & End \\
\midrule
\datainput{olmo_correct_rows.tex}
\bottomrule
\end{tabular}
\end{center}

\clearpage
\captionof{table}{\textbf{OLMo answer transitions for endpoint-donor grafts.} Correct, fixed and broken are counts out of the same 87 problems, listed for the first, second and third generation runs. Gain is the mean accuracy change from the paired ungrafted host, in percentage points. Retention and recovery are averages of run-level percentages. The retention denominators are the 30/39/35 initially correct problems; recovery denominators are the 57/48/52 initially incorrect problems. Fixed counts incorrect-to-correct transitions and broken counts correct-to-incorrect transitions.}
\label{tab:si_olmo_transitions}
\begin{center}\small\setlength{\tabcolsep}{5pt}
\begin{tabular}{llrrrll}
\toprule
\rowcolor{gray!15}Layers & Correct & Gain (pp) & Retention (\%) & Recovery (\%) & Fixed & Broken \\
\midrule
\datainput{olmo_endpoint_rows.tex}
\bottomrule
\end{tabular}
\end{center}

\clearpage
\notetitle{Interpreting donor-likeness scores}
\textbf{What the two scores compare.}
Both scores are zero for the host and one for the donor (Methods).
The donor-trace score $R_G$ measures the fraction of host--donor KL divergence closed by grafting on shared donor-generated response prefixes, using each model's prompt template.
Its token sums give longer traces more weight, and negative values indicate greater divergence from the donor than for the host on those response prefixes.
The generated-solution score $\operatorname{pos}_G$ is computed on each graft's own solutions, so it compares responses that follow different prefixes, and it can lie outside the interval from zero to one.
It measures relative likelihood under donor and host and does not measure agreement of final answers or of solution strategy.
Because the two scores use different token sets, they need not move together and their magnitudes are not directly comparable.

\textbf{Values.}
In Supplementary Table~\ref{tab:si_policy}, the region graft exceeded the mean of the non-overlapping grafts in 15 of 18 host--benchmark cells for each score; the three exceptions were M8B in both cases.
The two scores differed in sign for the M14B region graft: donor-trace scores were negative on all three benchmarks ($-0.16$ to $-0.15$), while generated-solution scores were positive (0.29 to 0.32).

\textbf{Sampling.}
The trace score uses one donor response per problem, and trace $n$ in Supplementary Table~\ref{tab:si_policy}b is the number of donor solutions scored.
Donor traces were generated with the mathematical instruction in the system turn, which for the Ministral reasoning models replaces the template's default instruction to produce a delimited thinking draft.
The generated-solution score uses the first response of the first generation run for each host and graft.
Class values weight positions equally, and the full-overlap class contains one region graft per host.

\captionof{table}{\textbf{Numerical donor-likeness summaries by overlap class.}
Overlap is the fraction of region layers included in a graft: None, $0$; Low, $(0,0.5)$; High, $[0.5,1)$; Full, $1$ (region graft). Both scores are zero for the ungrafted host and one for the donor. \textbf{a}, Number of distinct evaluated graft positions. \textbf{b}, Donor-trace score $R_G$, averaged across positions within each class. Trace $n$ is the number of retained donor problems (out of 87 for MATH-87 or 30 for each AIME year); over-limit exclusion and truncation follow Methods. \textbf{c}, Generated-solution score, averaged across the same positions. These are the values summarized graphically in Extended Data Fig.~5.}
\label{tab:si_policy}
\begin{center}\small
\begin{tabular}{lrrrrr}
\multicolumn{6}{l}{\textbf{a}\quad Evaluated positions}\\[3pt]
\toprule
\rowcolor{gray!15}Host & None & Low & High & Full & Total \\
\midrule
\datainput{policy_count_rows.tex}
\bottomrule
\end{tabular}

\end{center}

\clearpage
{\small\bfseries Supplementary Table~\ref{tab:si_policy} (continued). Numerical donor-likeness summaries by overlap class.}
\begin{center}\small
\begin{tabular}{llrrrrr}
\multicolumn{7}{l}{\textbf{b}\quad Donor-trace score}\\[3pt]
\toprule
\rowcolor{gray!15}Host & Benchmark & Trace $n$ & None & Low & High & Full \\
\midrule
\datainput{policy_R_rows.tex}
\bottomrule
\end{tabular}
\end{center}

\vspace{6pt}
\begin{center}\small
\begin{tabular}{llrrrr}
\multicolumn{6}{l}{\textbf{c}\quad Generated-solution score}\\[3pt]
\toprule
\rowcolor{gray!15}Host & Benchmark & None & Low & High & Full \\
\midrule
\datainput{policy_pos_rows_no_n.tex}
\bottomrule
\end{tabular}
\end{center}